\documentclass[11pt]{article}

\usepackage[preprint]{acl}
\usepackage{times}
\usepackage{latexsym}

\usepackage[T1]{fontenc}

\usepackage[utf8]{inputenc}

\usepackage{microtype}

\usepackage{inconsolata}
\usepackage{hyperref}
\usepackage{url}

\usepackage{booktabs}       
\usepackage{amsfonts}       
\usepackage{nicefrac}       
\usepackage{microtype}      
\usepackage{xcolor}         
\usepackage{pifont}         

\usepackage{graphicx}
\usepackage{subcaption}

\usepackage{amsmath}
\usepackage{amssymb}
\usepackage{mathtools}
\usepackage{amsthm}
\usepackage{xspace}

\usepackage{algorithm}
\usepackage{algorithmic}
\usepackage{enumitem}
\usepackage{multirow}
\usepackage{array}
\usepackage{makecell}

\usepackage[capitalize,noabbrev]{cleveref}

\theoremstyle{plain}
\newtheorem{theorem}{Theorem}[section]

\newtheorem{lemma}[theorem]{Lemma}

\theoremstyle{definition}

\theoremstyle{remark}

\usepackage{needspace}

\usepackage{listings}

\newcommand{\method}{\textbf{\texttt{OMCE}}\xspace}

\title{Evolving Memory, Learning to Meta-Control:\\ Test-Time Adaptation for Experience-augmented LLM Agents}

\author{
  \textbf{Zeyuan Zhao\textsuperscript{1,*}},
  \textbf{Yuanyuan Shi\textsuperscript{1,*}},
  \textbf{Peng Yu\textsuperscript{1}},
  \textbf{Lingyue Fu\textsuperscript{1}}, \\
  \textbf{Shao Zhang\textsuperscript{2,$\dagger$}},
  \textbf{Ying Wen\textsuperscript{1,$\dagger$}}\\
  {\normalfont
  \textsuperscript{1}Shanghai Jiao Tong University},
  {\normalfont
  \textsuperscript{2}Shanghai Artificial Intelligence Laboratory}\\
  {\normalfont\small
  \texttt{\{zhaozeyuan1102,syysyysyy1010,pursuit\_yp,fulingyue,ying.wen\}@sjtu.edu.cn}}\\
  {\normalfont\small
  \texttt{zhangshao@pjlab.org.cn}}
}

\begin{document}
\maketitle

\begingroup
\renewcommand{\thefootnote}{}
\footnotetext{%
  \begin{tabular}{@{}l@{}}
    \textsuperscript{*}Equal contribution. \\
    \textsuperscript{$\dagger$}Corresponding authors.
  \end{tabular}
}
\endgroup

\begin{abstract}

Self-evolving agents accumulate experience from task streams, but existing frameworks typically rely on fixed pipelines to retrieve and utilize this experience, neglecting when and how to deploy it. We introduce \textbf{Online Meta-Control of Experience} (\method), a framework that updates experience memory and learns a meta-control policy from task feedback during test-time adaptation.
\method employs a lightweight contextual-bandit meta-controller that uses compact retrieval signals to select among four cognitive actions---DIRECT, EPISODE, GUIDE, and DISTILL. This routing mechanism operationalizes three coupled decisions: whether to invoke experience, which memory format to utilize, and when to consolidate new guidance. \method evolves online, leveraging task feedback to update the memory bank and refine the routing policy via contextual bandit updates.
Experiments on four online adaptation benchmarks across four backbones show that \method improves first-attempt success by up to 20.5 percentage points over the strongest compared baseline, while reducing per-task token use by a factor of 5.7--62.5 relative to ACE with GPT-5.2.
Controlled ablations further support the benefits of learned routing and continued routing policy adaptation.
\end{abstract}

\section{Introduction}

Experience-oriented self-evolving LLM agents turn interaction feedback into reusable experience for later tasks \cite{gao2025survey, fang2025comprehensive, chen2026recursive}.
This experience can preserve concrete task cases or take more abstract forms, including reasoning guidance, reusable skills, and context playbooks \cite{fang2025memp, ouyang2025reasoningbank, wang2023voyager, zhang2025ACE}.
Its availability, however, does not guarantee its utility for the current task: semantic similarity alone may retrieve unhelpful experience \cite{zhang2026memrl}.
Even when experience is useful, concrete cases retain execution details but may not transfer readily to different tasks, while distilled guidance captures reusable principles but can omit details needed for execution \cite{fang2025memp, ouyang2025reasoningbank}.
Extending this memory also poses a further choice: distilling guidance from accumulated episodes may help future tasks, but also incurs additional computational cost and can lead to excessive memory growth\cite{zhang2025ACE}.
These trade-offs leave open how experience should be used and expanded as tasks arrive.

\begin{figure*}[t]
    \centering

    \includegraphics[width=0.96\textwidth]{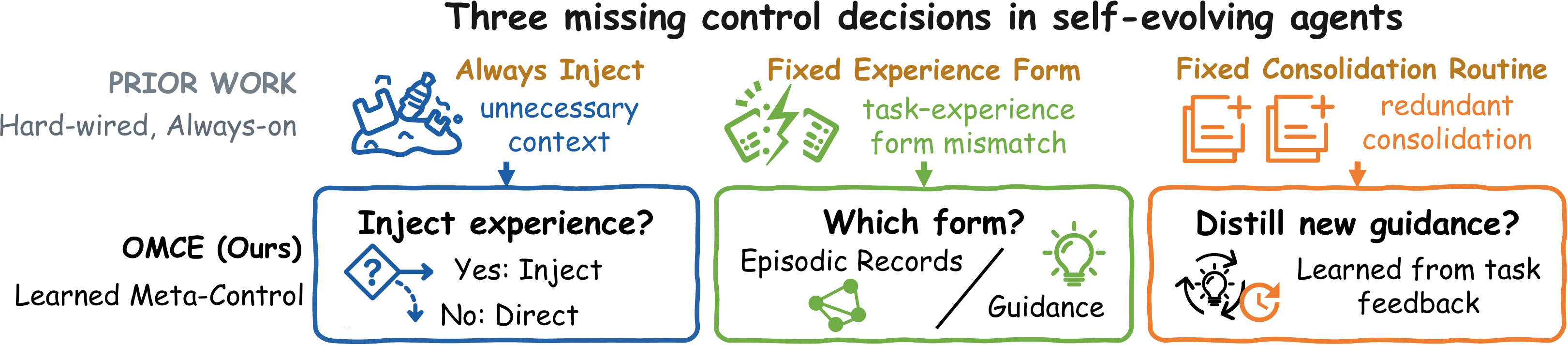}
    \vspace{-5pt}
    \caption{\textbf{From Fixed Pipelines to Learned Meta-Control. } \textbf{Up:} Existing methods typically follow a fixed routine for using and consolidating experience, even when the retrieved experience does not help the current task. \textbf{Down:} \method learns from task feedback when to use experience, whether to draw on past cases or general guidance, and when to distill new guidance.}
    \label{pic:Compared_with_prior}
\end{figure*}

We argue that scheduling accumulated experience is a learnable dimension of self-evolution, and make it precise through three requirements.
(1) \emph{Adaptive invocation}: whether experience is needed at all \cite{fang2025memp, ouyang2025reasoningbank}.
(2) \emph{Selective form}: which of its forms to use \cite{fang2025memp, zhang2025ACE}.

(3) \emph{Learnable consolidation}: when to distill transferable guidance from episodic evidence \cite{zhang2025ACE, shinn2023reflexion}.

Meta-control research on adaptive tool use \cite{li2025AdaptiveToolUse} and cognitive-demand monitoring \cite{scoville2026cognitive} suggests that adapting reasoning effort to task demands can be more effective than uniformly allocating computation \cite{de2024RationalMetareasoning}.These frameworks, however, control external, fixed resources such as a tool set or a reasoning budget. Accumulated experience differs: the agent produces it itself, in heterogeneous forms, in a store that grows and shifts, so scheduling it requires a decision over use \emph{and} consolidation, not a single invocation flag.

To this end, we propose \textbf{Online Meta-Control of Experience} (\method), an agent framework that learns to control experience use and consolidation through a lightweight meta-controller over episodic and guidance memory.

Implemented as a contextual bandit, the controller regulates whether to invoke experience, which form to use, and when to distill new guidance (Fig.~\ref{pic:Compared_with_prior}).

At inference, it summarizes retrieval profiles in a nine-dimensional meta-state vector and uses linear reward models to select among DIRECT, EPISODE, GUIDE, and DISTILL, without an additional LLM call for action selection.

After execution, memory evolution records outcomes, generates case-specific insights, and refines injected guidance using task feedback, while meta-evolution adapts the routing policy through online bandit updates.

Experiments on four online adaptation benchmarks show that \method achieves the highest first-attempt success in most evaluated settings, with competitive cumulative success under the same retry budget.
Ablations show that no fixed experience strategy is consistently the best, while controlled comparisons with shared memory demonstrate the value of learning \emph{when} and \emph{how} to use experience.
Freezing the controller reduces first-attempt success, supporting continued policy adaptation as memory evolves.
Finally, efficiency analysis shows that \method achieves a favorable performance--token trade-off, using substantially fewer tokens than ACE's full-playbook prompting.

\section{Related Work}

\paragraph{Experience Memory and Self-Evolving Agents.}

Experience-augmented reasoning has advanced LLM-agent memory systems from case-level trajectory retrieval \cite{fang2025memp,shinn2023reflexion} to structured agentic knowledge bases \cite{tang2025AgentKB,xu2025AgenticMem}.
Other work stores reusable plans, skills, or heuristics for cross-task transfer \cite{wang2023voyager,wu2402copilot,chen2025SweExp,allard2026experiential}, or distills experience into general principles, evolving playbooks, and reusable reasoning memories through generation, reflection, and refinement \cite{zhao2024expel,zhang2025ACE,ouyang2025reasoningbank}.
MemRL \cite{zhang2026memrl} improves retrieval by scoring memory candidates with a non-parametric Q-value function, and Evo-Memory \cite{wei2025evo} benchmarks retrieval, integration, and update in continuous task streams.
Overall, these efforts show that experience can be accumulated, abstracted, and maintained in multiple forms.
However, most frameworks still use a fixed retrieve-inject-reflect pipeline, with limited control over whether memory is useful for the current task, which type of experience to consult, or whether a new outcome should be consolidated \cite{majumder2023clin,yang2025MUSE}.

\paragraph{Meta-Control in LLM Agents.}

Learned meta-control decides whether to invoke a resource and how much to spend on it: a tool \cite{li2025AdaptiveToolUse}, a deliberation budget \cite{de2024RationalMetareasoning,ji2025MetaCognitiveMonitoring}, test-time compute \cite{snell2024scaling}, a learning mode \cite{dupoux2026ai}, or a verifier \cite{wan2026inference}; related work also edits the agent's program directly \cite{zhang2026hyperagents}.
Dual-process agents separate fast execution from slower deliberation \cite{evans2011dual,evans2013dual,liu2023llm,zhang2025DPTAgent}, and feedback-driven methods improve these meta-level policies with reward signals \cite{wu2025MetaOptimization}.
Closest to our setting are controllers selecting among reasoning strategies at inference time: Meta-Reasoner \cite{sui2026meta} learns with a contextual bandit when to backtrack or restart within one problem, and MetaScale \cite{liu2026metascale} evolves meta-thoughts as the sampling budget grows.
Both focus on controlling predefined resources, but \method instead learns to manage the experience an agent accumulates over time, jointly deciding when to use existing experience and when to distill new guidance.

\begin{figure*}[ht]
    \centering

    \includegraphics[width=0.96\textwidth]{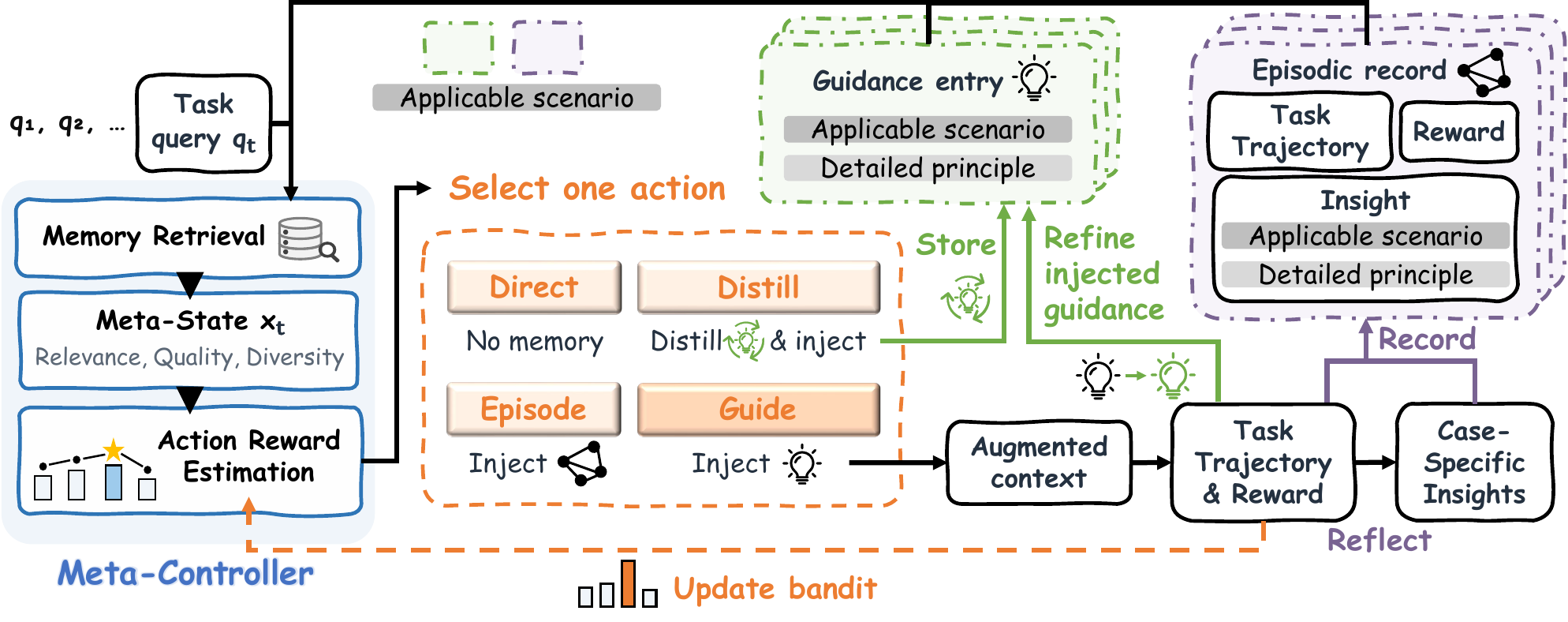}

    \caption{\textbf{Architecture of \method.} The lightweight Meta-Controller selects a cognitive action specifying whether and which experience to inject into the task prompt. The agent solves the task with the augmented context containing the query and any injected experience. Memory evolution records outcomes in episodic memory, distills guidance before execution when DISTILL is selected, and refines injected guidance afterward using task feedback.}
    \label{pic:ArchitectureOverview}
\end{figure*}

\section{Online Meta-Control of Experience}

Given a sequential stream of queries $\mathcal{Q}=\{q_1,\ldots,q_N\}$, \method equips an LLM agent with Episodic Memory, Guidance Memory, and a Meta-Controller (Fig.~\ref{pic:ArchitectureOverview}).

For each query, the controller uses meta-state to select an action that determines whether and which experience to inject into the task prompt, including whether to distill new guidance before execution. The resulting augmented context is supplied to the agent.
Task feedback then updates the stored experience and the controller's policy, allowing both to adapt across the stream.

\subsection{Memory Representation}
\label{sec:FrameworkArchitecture}

The \textbf{Episodic Memory} maintains episodic records of past interactions, $\mathcal{M}_\mathrm{epi}=\{\eta_i\}_{i=1}^{N}$.
Each record $\eta_i = (\tau_i, r_i, \Omega_i)$ contains the task trajectory $\tau_i$, the immediate reward $r_i \in [0,1]$, and a small set of case-specific insight entries $\Omega_i=\{\omega_{i,m}\}_{m=1}^{M_i}$.
These entries capture reusable patterns from successful cases or corrective lessons from failed ones.
The \textbf{Guidance Memory} stores guidance entries distilled from related episodic records, $\mathcal{M}_\mathrm{guide}=\{\psi_k\}_{k=1}^{K}$.
Compared with $\omega_{i,m}$, each $\psi_k$ is less tied to a single trajectory and instead summarizes a recurring principle that may transfer across similar cases.
For retrieval, both case-specific insights and guidance entries use the same two-part format: a compact description of the applicable scenario and a detailed principle.
The description serves as the cue for query-to-memory matching, while the principle supplies the content for injection into the task prompt.

\subsection{Adaptive Meta-Control}
\label{sec:AdaptiveMetaControl}

We model the inference-time control process of \method as a contextual bandit \cite{chu2011contextual,abbasi2011improved}, in which the \textbf{Meta-Controller} observes the meta-state $S_t^{\mathrm{meta}}$, selects a cognitive action, and receives feedback only for that action at each step.

\paragraph{Meta-State Representation.}

To support efficient action selection, the Meta-Controller summarizes the current retrieval results rather than consuming raw memory text.
Given the current query $q_t$, we retrieve the top-$M$ candidate entries from each memory source based on cosine similarity between the semantic embeddings of the query and each entry's applicable scenario \cite{reimers2019sentence,karpukhin2020dense}. We denote the retrieved sets from episodic memory and guidance memory by $\mathcal{M}^{\mathrm{retr}}_{\mathrm{epi}}$ and $\mathcal{M}^{\mathrm{retr}}_{\mathrm{guide}}$, respectively.
The meta-state is a nine-dimensional feature vector

\begingroup
\setlength{\abovedisplayskip}{1pt}
\setlength{\belowdisplayskip}{1pt}
\setlength{\abovedisplayshortskip}{1pt}
\setlength{\belowdisplayshortskip}{1pt}
\begin{equation}
\label{eq:meta_state}
S_t^{\mathrm{meta}} \equiv x_t=(x_{t,1},\ldots,x_{t,8},1).
\end{equation}
\endgroup

The state captures three properties: \emph{Relevance} ($x_{t,1},x_{t,2}$), measured by retrieval scores from the two memory sources; \emph{Quality} ($x_{t,3},\ldots,x_{t,6}$), measured by memory quality scores and their relevance-weighted counterparts; and \emph{Diversity} ($x_{t,7},x_{t,8}$), defined specifically as \emph{episodic outcome diversity}.
Diversity captures the balance between successful and failure-side episodic evidence and the normalized entropy of outcome types, rather than semantic diversity among retrieved entries.
The first eight coordinates are normalized online, while the final coordinate is a constant bias; $x_{t,7}$ and $x_{t,8}$ are exposed only to the DISTILL action through an action-specific feature mask. \Cref{app:meta_state} gives the full definitions and the quality-score mapping.

These three properties inform action selection. High relevance and reliable evidence favor experience reuse, weak or conflicting signals favor DIRECT, and relevant but insufficiently generalized episodes favor DISTILL.

\paragraph{Actions and Augmented Context.}
Given $x_t$, the Meta-Controller selects $a_t$ from $\mathcal{A}=\{a_{\textsc{DIRECT}},a_{\textsc{EPISODE}},a_{\textsc{GUIDE}},a_{\textsc{DISTILL}}\}$.
The selected action determines whether and which experience to inject into the task prompt containing $q_t$, producing the complete execution context $\mathcal{C}_{\mathrm{aug},t}$.
$a_{\textsc{DIRECT}}$ (\emph{DIRECT}, direct execution) retains the original task input without adding external experience.
$a_{\textsc{EPISODE}}$ (\emph{EPISODE}, episodic reuse) injects retrieved episodic records into the task prompt as concrete prior cases.
$a_{\textsc{GUIDE}}$ (\emph{GUIDE}, guidance reuse) injects retrieved guidance entries into the task prompt as reusable principles.
$a_{\textsc{DISTILL}}$ (\emph{DISTILL}, experience distillation) first distills and stores a new guidance entry $\psi_{\mathrm{new}}$ from the retrieved episodic evidence (\Cref{sec:MechanismI}), then injects it into the task prompt:
\begin{equation}
\label{eq:augmented_context}
\mathcal{C}_{\mathrm{aug},t}=
\begin{cases}
q_t, & a_t=a_{\textsc{DIRECT}},\\
\operatorname{Augment}(q_t,\mathcal{M}_{\mathrm{epi}}^{\mathrm{retr}}),
& a_t=a_{\textsc{EPISODE}},\\
\operatorname{Augment}(q_t,\mathcal{M}_{\mathrm{guide}}^{\mathrm{retr}}),
& a_t=a_{\textsc{GUIDE}},\\
\operatorname{Augment}(q_t,\{\psi_{\mathrm{new}}\}),
& a_t=a_{\textsc{DISTILL}}.
\end{cases}
\end{equation}
Here, $\operatorname{Augment}$ injects the selected experience into the task prompt containing the query, following the task prompt template without an additional LLM call. Fixed instructions shared by all actions are implicit, so the DIRECT branch denotes the original task input with those instructions.

\paragraph{Action Selection and Execution.}
Under the contextual-bandit formulation, the Meta-Controller maximizes the immediate current-query reward.
For each $a\in\mathcal{A}$, we use an action-wise linear model \cite{chu2011contextual,abbasi2011improved} $\mu_a(x_t)=x_t^\top\theta_a^\star$ with online estimate $\hat{r}_a^{(t)}(x_t)=x_t^\top\hat{\theta}_a^{(t)}$.
The estimate and decision rule are
\begingroup
\setlength{\abovedisplayskip}{2pt}
\setlength{\belowdisplayskip}{2pt}
\setlength{\abovedisplayshortskip}{1.5pt}
\setlength{\belowdisplayshortskip}{2pt}
\begin{equation}
\label{eq:theta_estimation}
\hat{\theta}^{(t)}_a = \left(A^{(t)}_a\right)^{-1} b^{(t)}_a, \quad
a_t = \operatorname*{arg\,max}_{a \in \mathcal{A}} \hat{u}_a^{(t)}(x_t).
\end{equation}
\endgroup
where $\hat{u}_a^{(t)}(x_t)$ combines the predicted reward with an exploration term, keeping underexplored actions selectable. The action-wise statistics $A_a^{(t)}$ and $b_a^{(t)}$ are updated from feedback as described below.
The agent executes the task conditioned on the augmented context, inducing a pretrained-LLM policy $\pi_\textsc{LLM}$ from which the task trajectory is sampled: $\tau_t \sim \pi_\textsc{LLM}(\cdot \mid \mathcal{C}_{\mathrm{aug},t})$.

\paragraph{Meta-Evolution.}
\label{sec:MechanismII}
Meta-evolution adapts the Meta-Controller through task feedback by updating the contextual bandit's action-wise reward models, which guide subsequent action selection. After execution, the agent receives reward $r_t$ for the current query. Only the selected action yields an observation, so the rewards of unchosen actions remain unobserved counterfactuals.

To update the bandit reward models, we update their action-wise sufficient statistics using $r_t$ as
\begingroup
\setlength{\abovedisplayskip}{3pt}
\setlength{\belowdisplayskip}{3pt}
\setlength{\abovedisplayshortskip}{2pt}
\setlength{\belowdisplayshortskip}{2pt}
\begin{equation}
\label{eq:stat_updated}
\begin{aligned}
A_a^{(t+1)} &= \gamma A_a^{(t)} + \mathbf{1}\{a_t=a\}x_t x_t^\top, \\
b_a^{(t+1)} &= \gamma b_a^{(t)} + \mathbf{1}\{a_t=a\}r_t x_t,
\quad \forall a\in\mathcal{A},
\end{aligned}
\end{equation}
\endgroup
where $0 < \gamma \le 1$ is a forgetting factor.
The new observation is added only to the selected action, while the discount is applied to all action models \cite{russac2019weighted} so that older evidence is downweighted under changing memory distributions.
The updated statistics determine the next parameter estimates through Eq.~\ref{eq:theta_estimation}. With four actions and a nine-dimensional meta-state, both action selection and bandit updates operate on small matrices and vectors, requiring no additional LLM calls.

A formal finite-time analysis of the controller's learning behavior is provided in Appendix~\ref{app:theorem}.

\subsection{Memory Evolution}
\label{sec:EvolutionaryMechanisms}
\label{sec:MechanismI}

\paragraph{DISTILL.}
Before execution, if $a_t=a_{\textsc{DISTILL}}$, an LLM distills a transferable principle from the retrieved episodic records and stores it as a guidance entry for current and future use:
\begingroup
\setlength{\abovedisplayskip}{2pt}
\setlength{\belowdisplayskip}{2pt}
\setlength{\abovedisplayshortskip}{2pt}
\setlength{\belowdisplayshortskip}{2pt}
\begin{equation}
\begin{aligned}
\psi_{\mathrm{new}} \leftarrow \operatorname{DISTILL}(\mathcal{M}_{\mathrm{epi}}^{\mathrm{retr}}),\\
\mathcal{M}_{\mathrm{guide}} \leftarrow \mathcal{M}_{\mathrm{guide}}\cup\{\psi_{\mathrm{new}}\}.
\end{aligned}
\label{eq:inductive_update}
\end{equation}
\endgroup

\paragraph{Episodic Record generation.}
After the agent produces a trajectory $\tau_t$ and receives reward $r_t$, an LLM extracts case-specific insights $\Omega_t=\{\omega_{t,m}\}_{m=1}^{M_t}$.
We record each execution outcome in Episodic Memory $\mathcal{M}_{\mathrm{epi}}$ as $\eta_{\mathrm{new}}=(\tau_t,r_t,\Omega_t)$, regardless of the selected action or task outcome.
This preserves evidence for later reuse and distillation, while the controller determines when to distill it into transferable guidance.

\paragraph{Guidance refinement.}
Refinement is a follow-up update to the guidance entries which are injected into the augmented context. After execution under $a_{\textsc{GUIDE}}$ or $a_{\textsc{DISTILL}}$, an LLM refines each such entry, whether retrieved or newly distilled, with the realized trajectory $\tau_t$ and reward $r_t$:

\begingroup
\setlength{\abovedisplayskip}{1pt}
\setlength{\belowdisplayskip}{2pt}
\setlength{\abovedisplayshortskip}{0.5pt}
\setlength{\belowdisplayshortskip}{2pt}
\begin{equation}
\begin{aligned}
\psi_k' &\leftarrow \operatorname{Refine}(\psi_k, \tau_t, r_t), \\ 
\mathcal{M}_{\mathrm{guide}} &\leftarrow \bigl(\mathcal{M}_{\mathrm{guide}} \setminus \{\psi_k\}\bigr) \cup \{\psi_k'\}.
\end{aligned}
\label{eq:deductive_update}
\end{equation}
\endgroup

\paragraph{Optional Multi-Attempt Extension.}
With additional inference budget, the agent may make up to $P$ attempts per task, stopping upon success \cite{snell2024scaling,shinn2023reflexion}.
After a failed attempt, the extracted case-specific experience is stored in Episodic Memory and injected into the task prompt containing the original query to form the augmented context for the next attempt on the same task.
Appendix~\ref{algo:omce} gives the complete inference and learning loop, including this extension.

\begin{table*}[htbp]
\centering
\scriptsize
\setlength{\tabcolsep}{4pt}
\renewcommand{\arraystretch}{1.15}
\resizebox{\textwidth}{!}{%
\begin{tabular}{>{\centering\arraybackslash}m{1.46cm}
                >{\centering\arraybackslash}m{2.0cm}
                *{1}{>{\centering\arraybackslash}m{1.76cm}}
                *{2}{>{\centering\arraybackslash}m{1.1cm}}
                *{2}{>{\centering\arraybackslash}m{1.82cm}}
                *{1}{>{\centering\arraybackslash}m{1.77cm}}
                *{1}{>{\centering\arraybackslash}m{1.77cm}}
                *{1}{>{\centering\arraybackslash}m{1.77cm}}}
\toprule
\multirow{3}{*}{\centering\textbf{Benchmark}} & \multirow{3}{*}{\centering\textbf{Model}} & \multicolumn{7}{c}{\textbf{OSR $\uparrow$ / CSR $\uparrow$}} \\
\cline{3-9}
 & & \textbf{Base LLM} & \textbf{Reflexion} & \textbf{MemP} & \textbf{ReasoningBank} & \textbf{ACE} & \method-TS  & \method-UCB \\
\midrule

\multirow{4}{*}{\centering\makecell{\textbf{Multi-}\\ \textbf{Challenge}}}
& GPT-5.2 & 52.8\% / 70.1\% & - / \textbf{98.2\%} & 50.2\% / - & 53.9\% / 66.3\% & 48.7\% / 71.4\% & \textbf{67.4\%} / 97.1\% & 66.3\% / 95.6\% \\
& Deepseek-V3.2 & 43.1\% / 55.7\% & - / 91.6\% & 46.9\% / - & 48.4\% / 56.0\% & 50.2\% / 70.0\% & \textbf{70.7\%} / 95.6\% & 67.4\% / \textbf{96.7\%} \\
& Deepseek-V4-pro & 49.5\% / 63.0\%  & - / 90.1\%  & 48.0\% / - & 55.3\% / 66.3\% & 50.2\% / 74.7\%  & \textbf{69.6\%} / \textbf{93.8\%} & 64.8\% / 91.2\% \\
& Kimi-K2.5 &  49.8\% / 64.8\% & - / 89.0\% &  50.9\% / - &  56.7\% / 72.8\% & 49.8\% / 70.7\%  & 72.2\% / 95.6\% & \textbf{73.3\%} / \textbf{96.7\%} \\
\midrule
\multirow{4}{*}{\centering\textbf{BBEH}}
& GPT-5.2          & 33.3\% / 49.7\% & - / 83.9\% & 31.9\% / - & 31.3\% / 37.7\% & 52.4\% / \textbf{94.4\%} & \textbf{53.3\%} / 78.7\% & 46.8\% / 71.0\% \\
& Deepseek-V3.2    & 41.7\% / 59.5\% & - / 88.7\% & 26.9\% / - & 37.3\% / 44.6\% & 45.2\% / \textbf{91.2\%} & 45.8\% / 73.8\% & \textbf{46.3\%} / 73.3\% \\
& Deepseek-V4-pro  & 38.3\% / 60.1\% & - / \textbf{93.3\%} & 33.8\% / - & \textbf{47.4\%} / 55.5\% & 45.7\% / 93.1\% & 47.3\% / 79.8\% & \textbf{47.4\%} / 80.0\% \\
& Kimi-K2.5 & 39.7\% / 61.5\% & - / 85.8\% & \textbf{48.8\%} / - & 15.6\% / 16.6\% & 46.2\% / \textbf{87.1\%} & 48.7\% / 80.7\% & 48.4\% / 81.2\% \\

\midrule

\multirow{4}{*}{\centering\makecell{\textbf{BigCode-}\\ \textbf{Bench}}}
& GPT-5.2 & 48.1\% / 54.0\% & - / 61.4\% & 51.1\% / - & 50.4\% / \textbf{68.1\%} & 38.2\% / 40.3\% & 49.6\% / 58.9\% & \textbf{51.3\%} / 59.2\% \\
& Deepseek-V3.2    & 47.9\% / 58.4\% & - / 69.0\% & \textbf{49.3\%} / - & 47.6\% / \textbf{69.7\%}          & 36.9\% / 41.0\% & 47.7\% / 65.2\% & 47.1\% / 65.5\% \\
& Deepseek-V4-pro  & 35.7\% / 52.1\% & - / 66.9\% & \textbf{49.6\%} / - & 35.9\% / 69.7\%          & 36.7\% / 41.4\% & 43.3\% / \textbf{70.7\%} & 44.1\% / 69.7\% \\
& Kimi-K2.5        & 49.1\% / 55.7\% & - / \textbf{73.4\%} & \textbf{51.1\%} / - & 49.3\% / \textbf{73.4\%}          & 38.0\% / 42.5\% & 48.5\% / 68.3\% & 49.1\% / 69.7\% \\
\midrule

\multirow{4}{*}{\centering\makecell{\textbf{SWE-bench} \\ \textbf{Verified}}}
& GPT-5.2 & 52.8\% / 64.0\% & - / 64.2\% & 61.8\% / - & 53.6\% / 63.0\% & 50.8\% / 52.0\% & \textbf{64.8\%} / \textbf{64.8\%} & \textbf{64.8\%} / \textbf{64.8\%} \\
& Deepseek-V3.2    & 40.2\% / 42.0\% & - / 46.6\% & 48.0\% / - & 46.8\% / 48.0\% & 44.2\% / 47.4\% & 52.4\% / 53.8\% & \textbf{53.0\%} / \textbf{54.0\%} \\
& Deepseek-V4-pro  & 48.2\% / 48.2\% & - / 53.4\% & 48.8\% / - & 42.6\% / 43.8\% & 43.8\% / 45.2\% & \textbf{58.8\%} / 60.4\% & 58.4\% / \textbf{61.2\%} \\
& Kimi-K2.5        & \textbf{49.2\%} / 50.2\% & - / \textbf{53.6\%} & 45.0\% / - & 47.0\% / 49.6\% & 45.2\% / 46.8\% & 48.8\% / 49.0\% & 48.6\% / 48.6\% \\
\bottomrule

\end{tabular}
}
\caption{Main results under the online adaptation setting. Each entry is reported as \textbf{OSR / CSR} (\%), where CSR is measured within $P=4$ attempts. For Reflexion, ``-'' denotes that no first-attempt evaluation applies, as explained in the metric definitions above.}
\label{tab:main_results}
\end{table*}

\section{Experiments}

We evaluate all methods in an online adaptation setting, where tasks arrive and must be solved sequentially. After completing an instance, the agent observes task feedback and may update its memory, controller, or context before processing the next instance.

\subsection{Experimental Design}

We organize the evaluation around five questions. \emph{RQ1}: does learned routing raise first-attempt success across benchmarks and backbones? \emph{RQ2}: how should the routing policy be determined, in particular is a fixed form of experience use sufficient, and does a learned controller beat fixed and hand-crafted alternatives? \emph{RQ3}: do the gains come from the routing policy, or from the memory content and the update schedule? \emph{RQ4}: does the controller keep improving as the stream proceeds? \emph{RQ5}: what is the token cost of these gains? \Cref{sec:main_results} addresses RQ1, \Cref{sec:ablation} addresses RQ2 to RQ4, and \Cref{sec:efficiency} addresses RQ5.

The evaluated methods span four test-time adaptation regimes: no external memory (\textbf{Base LLM}), within-instance retry (\textbf{Reflexion}~\cite{shinn2023reflexion}), cross-task adaptation (\textbf{MemP}~\cite{fang2025memp}), and cross-task adaptation with multi-attempt inference, including \textbf{ACE}~\cite{zhang2025ACE}, \textbf{ReasoningBank}~\cite{ouyang2025reasoningbank} and our \method, which differs from these baselines in learning when to invoke experience and which form to use rather than following a fixed routine. Against \method, we additionally compare policies that always commit to one form of experience and a hand-crafted rule over retrieval scores (\Cref{sec:ablation}). We report two variants of \method: \method-TS and \method-UCB, which implement meta-evolution with two contextual-bandit action-selection algorithms: Thompson Sampling (TS) \cite{agrawal2013thompson} and LinUCB \cite{chu2011contextual,abbasi2011improved}, respectively (detailed in Appendix~\ref{app:action_selection}).

We evaluate on four complementary benchmarks spanning distinct task domains under this shared setting: BigCodeBench \cite{zhuo2024bigcodebench} and SWE-bench Verified \cite{jimenez2023swe} for code generation and repository-level software engineering, MultiChallenge \cite{deshpande2025multichallenge} for realistic multi-turn conversation, and BBEH \cite{kazemi2025big} for advanced general reasoning.
Across all benchmarks, we use four LLMs as backbone models: GPT-5.2 \cite{web:openai:gpt52}, DeepSeek-V3.2 \cite{liu2025deepseek}, DeepSeek-V4-pro \cite{deepseekai2026deepseekv4}, and Kimi-K2.5 \cite{team2026kimi}.

We use \emph{one-shot success rate} (OSR) as the primary metric, also called first-attempt success: it measures whether the current policy and accumulated experience solve an instance on the first attempt.
For methods with retry mechanisms, we additionally report \emph{cumulative success rate} (CSR), defined as success within a shared budget of $P=4$ total attempts, including the initial attempt (\Cref{app:metrics}). Reflexion does not transfer experience across instances, so by construction its first attempt matches the Base LLM under this protocol; we therefore report its CSR only, while \method's CSR combines within-instance retries with cross-instance adaptation.
Full details on benchmark construction, trial protocol, baseline implementations, metrics, and controller variants are provided in Appendix~\ref{app:design}. Both variants share one configuration across all benchmarks and backbones; the values and the selection procedure are given there.

\begin{table*}[!t]
\centering
\footnotesize
\setlength{\tabcolsep}{3pt}
\renewcommand{\arraystretch}{1.03}
\captionsetup{skip=4pt}

\begin{tabular*}{\textwidth}{@{\extracolsep{\fill}}lccccc@{}}
\toprule
\multirow{2}{*}{\textbf{Benchmark}}
& \multicolumn{5}{c}{\textbf{OSR $\uparrow$ / CSR $\uparrow$}} \\
\cmidrule(lr){2-6}
& \textbf{Fixed-}$a_{\textsc{EPISODE}}$
& \textbf{Fixed-}$a_{\textsc{GUIDE}\&\textsc{DISTILL}}$
& \method-HEU
& \method-TS
& \method-UCB \\
\midrule

\textbf{MultiChallenge}
& 56.0\% / 91.2\%
& 63.7\% / 94.5\%
& 64.5\% / 95.2\%
& \textbf{67.4\%} / \textbf{97.1\%}
& 66.3\% / 95.6\% \\

\textbf{BBEH}
& 50.3\% / 76.8\%
& 43.1\% / 73.8\%
& 45.6\% / 71.7\%
& \textbf{53.3\%} / \textbf{78.7\%}
& 46.8\% / 71.0\% \\

\textbf{BigCodeBench}
& 49.0\% / 59.0\%
& 49.6\% / 58.0\%
& 47.4\% / 57.4\%
& 49.6\% / 58.9\%
& \textbf{51.3\%} / \textbf{59.2\%} \\

\textbf{SWE-bench Verified}
& 64.2\% / \textbf{64.8\%}
& 64.4\% / \textbf{64.8\%}
& 64.0\% / \textbf{64.8\%}
& \textbf{64.8\%} / \textbf{64.8\%}
& \textbf{64.8\%} / \textbf{64.8\%} \\

\bottomrule
\end{tabular*}

\caption{Routing ablation of \method on GPT-5.2, OSR / CSR (\%).
Fixed-$a_{\textsc{EPISODE}}$ and
Fixed-$a_{\textsc{GUIDE}\&\textsc{DISTILL}}$
always commit to one form of experience;
\method-HEU selects actions with hand-crafted rules.
Appendix~\ref{app:routing_ablation} reports the same comparison
for the remaining backbones.}
\label{tab:ablation}
\end{table*}

\begin{table*}[t]
\centering
\scriptsize
\renewcommand{\arraystretch}{1.15}
\resizebox{\textwidth}{!}{%
\begin{tabular}{llccccc}
\toprule
 & & & \multicolumn{2}{c}{\textbf{Online memory evolution}} & \multicolumn{2}{c}{\textbf{Shared memory snapshot}} \\
\cmidrule(lr){4-5} \cmidrule(lr){6-7}
\textbf{Benchmark} & \textbf{Variant} & \method & \textbf{Always-Both-and-Update} & \textbf{Frozen Controller} & \textbf{Always-Both} & \textbf{Random} \\
\midrule
MultiChallenge     & \method-TS  & \textbf{67.4\%} / 97.1\% & 63.7\% / 95.2\% & 61.9\% / 94.1\% & 63.0\% / 83.5\% & 62.6\% / 83.9\% \\
BBEH               & \method-TS  & \textbf{53.3\%} / 78.7\% & 51.1\% / 77.6\% & 49.8\% / 75.6\% & 49.7\% / 71.8\% & 48.2\% / 67.4\% \\
BigCodeBench       & \method-UCB & \textbf{51.3\%} / 59.2\% & 48.3\% / 53.5\% & 49.5\% / 65.4\% & 47.5\% / 55.6\% & 46.5\% / 57.0\% \\
SWE-bench Verified & \method-TS  & \textbf{64.8\%} / \textbf{64.8\%} & 55.0\% / 55.2\% & 64.2\% / 64.6\% & 48.8\% / 49.2\% & 53.0\% / 53.4\% \\
\bottomrule
\end{tabular}
}
\caption{Controlled analysis on GPT-5.2, OSR / CSR (\%). For each benchmark, \method uses the variant with the higher GPT-5.2 OSR in \Cref{tab:main_results} (TS on ties). Frozen Controller stops controller updates mid-stream while memory keeps evolving; Always-Both and Random read one shared memory snapshot without updating it.}
\label{tab:controlled}
\end{table*}

\subsection{Main Results}
\label{sec:main_results}

Table~\ref{tab:main_results} reports the performance across all benchmark-backbone configurations, demonstrating that \method systematically improves test-time adaptation: it attains the highest first-attempt success on eleven of the sixteen benchmark--backbone configurations. We highlight two key findings:

\textbf{Learned Routing Leads in First-Attempt Success.} \method dominates first-attempt success wherever retrieved experience carries reliable signal. On MultiChallenge it leads ReasoningBank, ACE, and MemP on all four backbones, by 13.5 to 20.5 points over the best of the three, and its weakest backbone there, at 67.4\%, still exceeds the strongest baseline on any backbone by 10.7 points. It leads the same three baselines on all four backbones of SWE-bench Verified by 1.8 to 10.0 points. On BBEH it matches or leads the strongest baseline on three backbones, reaching 53.3\% with GPT-5.2 against 52.4\% for ACE and trailing by 0.1 point on Kimi-K2.5. On BigCodeBench, \method leads in OSR with GPT-5.2 but trails MemP on the other three backbones.

\textbf{Selective Routing Can Bypass Memory.} DIRECT allows the controller to solve a task without injecting retrieved experience. On BigCodeBench, the better-performing \method variant matches or exceeds the Base LLM's OSR on three backbones and trails it by 0.2 point on the fourth. ACE falls 9.9 to 11.1 points below it on three.


\subsection{Ablation Study}
\label{sec:ablation}

We examine where the gains of \method come from through three questions, answering \emph{RQ2} to \emph{RQ4}. The first compares routing rules on all four backbones, covering two fixed policies, a hand-crafted one, and the two learned variants: \Cref{tab:ablation} reports GPT-5.2 and \Cref{app:routing_ablation} the full grid. The other two examine memory use and controller updates on GPT-5.2, holding the backbone fixed across conditions.

\begin{figure*}[t]
\centering
\includegraphics[width=\textwidth]{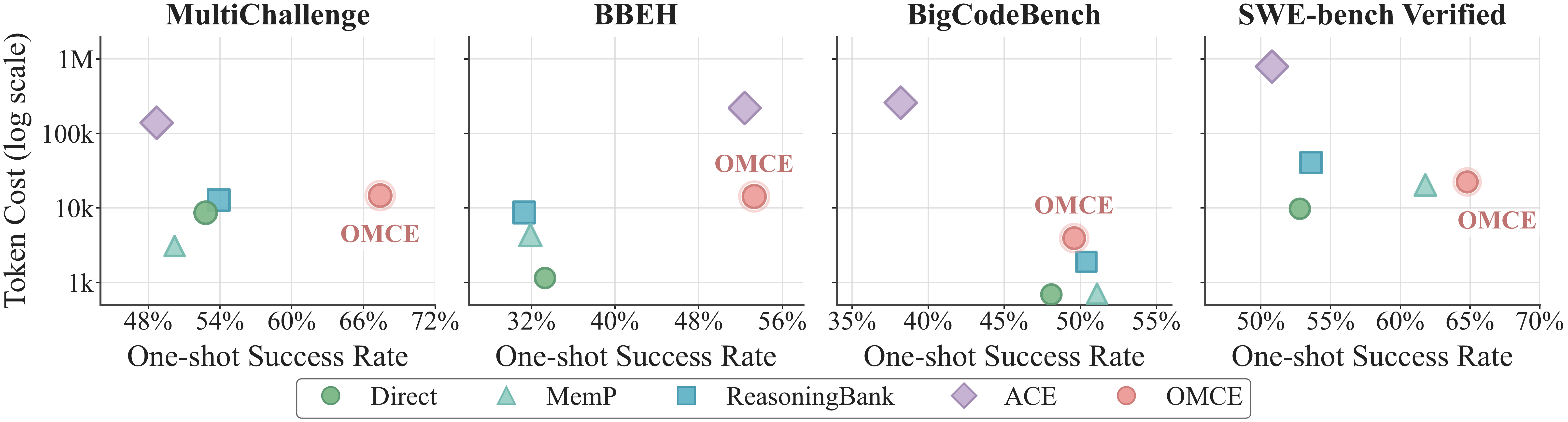}
\caption{\textbf{Performance--efficiency trade-off on four benchmarks} using GPT-5.2. The x-axis represents the one-shot success rate, and the y-axis reports the token cost in log scale (average total prompt and output tokens consumed per task).}
\label{pic:performance_vs_efficiency}
\end{figure*}

\paragraph{How should the routing policy be determined?}
Fixed-$a_{\textsc{EPISODE}}$ always injects retrieved episodic records into the task prompt, and Fixed-$a_{\textsc{GUIDE}\&\textsc{DISTILL}}$ always reuses or distills guidance. On GPT-5.2, Fixed-$a_{\textsc{EPISODE}}$ exceeds Fixed-$a_{\textsc{GUIDE}\&\textsc{DISTILL}}$ by 7.2 OSR points on BBEH (50.3 vs.\ 43.1), and the order reverses on MultiChallenge, where Fixed-$a_{\textsc{GUIDE}\&\textsc{DISTILL}}$ leads by 7.7 points (63.7 vs.\ 56.0). Committing to either form costs accuracy where it fits poorly: Fixed-$a_{\textsc{EPISODE}}$ trails \method-TS by 11.4 points on MultiChallenge, and Fixed-$a_{\textsc{GUIDE}\&\textsc{DISTILL}}$ trails it by 10.2 points on BBEH. Which form helps therefore depends on the task; the same reversal appears at the level of individual backbones, where the BBEH ordering of the two fixed policies flips between GPT-5.2 and DeepSeek-V3.2.

\method-HEU is also adaptive, selecting among the same four actions with fixed thresholds on retrieval scores (Appendix~\ref{app:heuristic_baseline}). On GPT-5.2 the learned controllers exceed the tuned heuristic by 0.8 to 7.7 points on all four benchmarks, with the widest margin on BBEH (53.3 vs.\ 45.6). Across the full grid the better learned variant outperforms \method-HEU on 14 of the 16 benchmark--backbone pairs.
Individually, TS and UCB each outperform \method-HEU in OSR on 11 of the 16 configurations.

\paragraph{Do the gains come from routing or from the memory?}
To separate routing from memory content, we fix the memory that \method has built and vary only the routing rule. \textit{Always-Both} reads both memory types on every task and \textit{Random} picks a feasible action uniformly, both against that same snapshot. With identical memory content, \method exceeds Always-Both by 3.6 to 16.0 OSR points and Random by 4.8 to 11.8, with higher CSR on all four benchmarks, and reaches the best OSR of any policy on the same snapshot. Where the gap is largest, on SWE-bench Verified, Always-Both reaches 48.8 OSR, below the Base LLM without any memory (52.8, \Cref{tab:main_results}), and on BigCodeBench it reaches 47.5 against 48.1. On two of the four benchmarks, then, indiscriminate retrieval is worse than retrieving nothing.

The same holds when the memory keeps evolving. \textit{Always-Both-and-Update} runs the full online loop as \method does, but retrieves from both memory types and updates memory after every task, and it trails \method on all four benchmarks by 2.2 to 9.8 OSR points, with lower CSR in each case. These comparisons show an OSR advantage for learned routing under both fixed and evolving memory.

\begin{figure*}[htbp]
    \vspace{-4pt}
    \centering
    \includegraphics[width=0.96\textwidth]{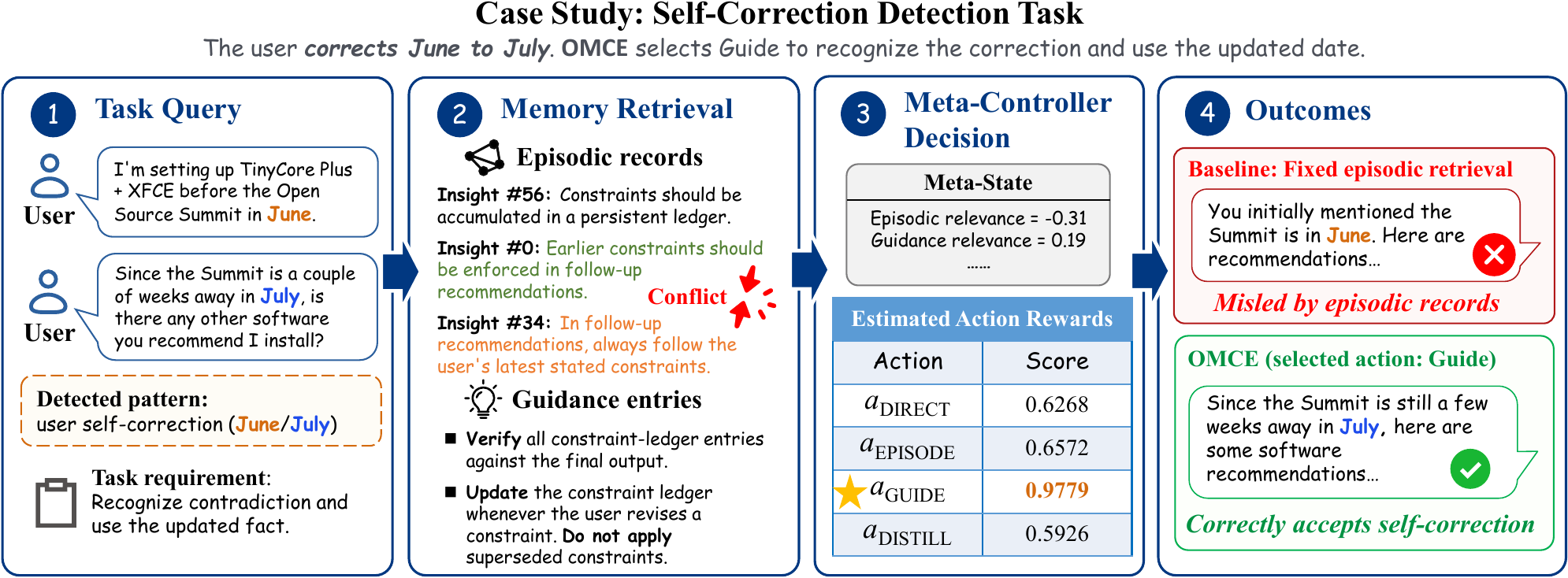}
    \caption{\textbf{Self-correction detection in MultiChallenge.} The meta-controller routes the query to GUIDE ($a_{\textsc{GUIDE}}$) rather than to episodic cases that enforce the user's outdated constraint.}
    \label{fig:case_study}
\end{figure*}

\paragraph{Does the controller need to keep learning?}
We freeze the controller parameters at the midpoint of each stream while memory continues to evolve. The two interventions in this section order the benchmarks identically. Freezing the controller costs first-attempt success on all four, by 5.5 points on MultiChallenge, 3.5 on BBEH, 1.8 on BigCodeBench, and 0.6 on SWE-bench Verified, while \method's lead over the best fixed policy of \Cref{tab:ablation} over the same four benchmarks is 3.7, 3.0, 1.7, and 0.4 points. CSR falls on three benchmarks, by 0.2 to 3.1 points; on BigCodeBench it instead rises, from 59.2 to 65.4. Continued controller updates improve OSR on all four benchmarks, while their effect on CSR varies across benchmarks.


\subsection{Efficiency Analysis: Token--Performance Trade-off}
\label{sec:efficiency}

We evaluate efficiency using \textit{tokens per task}, the average total LLM tokens consumed per instance over the full online adaptation pipeline; \Cref{app:metrics} gives the exact accounting.
\Cref{pic:performance_vs_efficiency} plots the performance--efficiency frontier with GPT-5.2 as the backbone; Appendix~\ref{app:efficiency_details} gives the statistics for all backbones.

\textbf{MemP} is the most token-efficient method because it performs no test-time scaling. With GPT-5.2, \method reduces per-task token use by a factor of 5.7--62.5 relative to \textbf{ACE}. \method sits between the two across all four benchmarks and all four backbone LLMs (\Cref{fig:efficiency_all_backbones}), at a token cost close to the retrieval baselines.

\subsection{Case Study}
A case from MultiChallenge illustrates the routing decision (\Cref{fig:case_study}): a user planning a conference initially says ``in June,'' then corrects it to ``in July.''

A fixed pipeline retrieves insights from the top-matching episodic records, all of which direct the agent to hold earlier statements as hard constraints, and the model accordingly insists on June.
The meta-controller reads $S^{\text{meta}}_{t}$ as weak or conflicting on the episodic side and routes the query to $a_{\textsc{GUIDE}}$, where the retrieved guidance concerns detecting and reconciling contradictions; the agent then picks up the July date and solves the task.

\section{Conclusion}

In this work, we presented \method, a framework that recasts experience-augmented reasoning from an always-on memory pipeline into a learnable meta-cognitive control problem. Instead of treating retrieval, context injection, reflection, and consolidation as a fixed routine, \method employs a lightweight meta-controller based on compact retrieval features and linear reward models to govern the full lifecycle of experience: choosing among DIRECT, EPISODE, GUIDE, and DISTILL. Crucially, this meta-controller evolves online via task feedback, progressively improving the agent's judgment over its own cognitive processes. Evaluations across four online adaptation benchmarks demonstrate that \method establishes state-of-the-art OSR and a favorable performance-token trade-off. Ultimately, our findings suggest that as self-evolving agents mature, the primary bottleneck of online adaptation shifts from merely accumulating richer experience to learning exactly when and how that experience should be deployed and updated.

\clearpage
\section*{Limitations}

\method inherently relies on timely feedback. In environments with sparse or delayed rewards, the contextual-bandit updates can become noisy, and abrupt task distribution shifts may render both the accumulated experience and the learned meta-policy stale. Moreover, due to the partial observability of the bandit formulation, the controller lacks counterfactual signals regarding unchosen cognitive routes, potentially slowing its recovery from early suboptimal decisions. Additionally, as the memory banks expand indefinitely, retrieval may suffer from redundancy or semantic collisions.

To address these limitations, future work could integrate established memory-maintenance techniques (e.g., explicit aging or eviction of stored memory entries). More fundamentally, future iterations could leverage the LLM's reflective capabilities to simulate counterfactual rewards for unchosen actions (potentially accelerating bandit convergence), or complement the current meta-control framework with explicit memory auditing and pruning, advancing towards a fully autonomous lifelong-learning agent.

Our controlled comparisons of memory use and controller updates focus on GPT-5.2, while the routing comparison spans all four backbones. Testing the same interventions on other backbones would help assess whether the observed effects generalize.

\section*{Acknowledgments}

We used AI assistants to polish and paraphrase author-written text, assist with short-form text completion, and identify relevant literature. All references were verified by the authors, who take full responsibility for the paper's content.

\bibliography{custom}

\begin{thebibliography}{56}
\providecommand{\natexlab}[1]{#1}

\bibitem[{Abbasi-Yadkori et~al.(2011)Abbasi-Yadkori, P{\'a}l, and Szepesv{\'a}ri}]{abbasi2011improved}
Yasin Abbasi-Yadkori, D{\'a}vid P{\'a}l, and Csaba Szepesv{\'a}ri. 2011.
\newblock Improved algorithms for linear stochastic bandits.
\newblock \emph{Advances in neural information processing systems}, 24.

\bibitem[{Agrawal and Goyal(2013)}]{agrawal2013thompson}
Shipra Agrawal and Navin Goyal. 2013.
\newblock Thompson sampling for contextual bandits with linear payoffs.
\newblock In \emph{International conference on machine learning}, pages 127--135. PMLR.

\bibitem[{Allard et~al.(2026)Allard, Teinturier, Xing, and Viaud}]{allard2026experiential}
Marc-Antoine Allard, Arnaud Teinturier, Victor Xing, and Gautier Viaud. 2026.
\newblock Experiential reflective learning for self-improving llm agents.
\newblock \emph{arXiv preprint arXiv:2603.24639}.

\bibitem[{Chen et~al.(2026)Chen, Wang, and Qu}]{chen2026recursive}
Mingguang Chen, Licheng Wang, and Bo~Qu. 2026.
\newblock Recursive self-improvement in ai: From bounded self-refinement to autonomous research loops.
\newblock \emph{arXiv preprint arXiv:2607.07663}.

\bibitem[{Chen et~al.(2025)Chen, Lin, Gu, Shi, Lian, Yun, Chen, Sun, Cao, and Wang}]{chen2025SweExp}
Silin Chen, Shaoxin Lin, Xiaodong Gu, Yuling Shi, Heng Lian, Longfei Yun, Dong Chen, Weiguo Sun, Lin Cao, and Qianxiang Wang. 2025.
\newblock Swe-exp: Experience-driven software issue resolution.
\newblock \emph{arXiv preprint arXiv:2507.23361}.

\bibitem[{Chu et~al.(2011)Chu, Li, Reyzin, and Schapire}]{chu2011contextual}
Wei Chu, Lihong Li, Lev Reyzin, and Robert Schapire. 2011.
\newblock Contextual bandits with linear payoff functions.
\newblock In \emph{Proceedings of the fourteenth international conference on artificial intelligence and statistics}, pages 208--214. JMLR Workshop and Conference Proceedings.

\bibitem[{De~Sabbata et~al.(2024)De~Sabbata, Sumers, AlKhamissi, Bosselut, and Griffiths}]{de2024RationalMetareasoning}
C~Nicol{\`o} De~Sabbata, Theodore~R Sumers, Badr AlKhamissi, Antoine Bosselut, and Thomas~L Griffiths. 2024.
\newblock Rational metareasoning for large language models.
\newblock \emph{arXiv preprint arXiv:2410.05563}.

\bibitem[{DeepSeek-AI(2026)}]{deepseekai2026deepseekv4}
DeepSeek-AI. 2026.
\newblock Deepseek-v4: Towards highly efficient million-token context intelligence.

\bibitem[{Deshpande et~al.(2025)Deshpande, Sirdeshmukh, Mols, Jin, Hernandez-Cardona, Lee, Kritz, Primack, Yue, and Xing}]{deshpande2025multichallenge}
Kaustubh Deshpande, Ved Sirdeshmukh, Johannes~Baptist Mols, Lifeng Jin, Ed-Yeremai Hernandez-Cardona, Dean Lee, Jeremy Kritz, Willow~E Primack, Summer Yue, and Chen Xing. 2025.
\newblock Multichallenge: A realistic multi-turn conversation evaluation benchmark challenging to frontier llms.
\newblock In \emph{Findings of the Association for Computational Linguistics: ACL 2025}, pages 18632--18702.

\bibitem[{Dupoux et~al.(2026)Dupoux, LeCun, and Malik}]{dupoux2026ai}
Emmanuel Dupoux, Yann LeCun, and Jitendra Malik. 2026.
\newblock Why ai systems don't learn and what to do about it: Lessons on autonomous learning from cognitive science.
\newblock \emph{arXiv preprint arXiv:2603.15381}.

\bibitem[{Evans(2011)}]{evans2011dual}
Jonathan St~BT Evans. 2011.
\newblock Dual-process theories of reasoning: Contemporary issues and developmental applications.
\newblock \emph{Developmental review}, 31(2-3):86--102.

\bibitem[{Evans and Stanovich(2013)}]{evans2013dual}
Jonathan St~BT Evans and Keith~E Stanovich. 2013.
\newblock Dual-process theories of higher cognition: Advancing the debate.
\newblock \emph{Perspectives on psychological science}, 8(3):223--241.

\bibitem[{Fang et~al.(2025{\natexlab{a}})Fang, Peng, Zhang, Wang, Yi, Zhang, Xu, Wu, Liu, Li et~al.}]{fang2025comprehensive}
Jinyuan Fang, Yanwen Peng, Xi~Zhang, Yingxu Wang, Xinhao Yi, Guibin Zhang, Yi~Xu, Bin Wu, Siwei Liu, Zihao Li, et~al. 2025{\natexlab{a}}.
\newblock A comprehensive survey of self-evolving ai agents: A new paradigm bridging foundation models and lifelong agentic systems.
\newblock \emph{arXiv preprint arXiv:2508.07407}.

\bibitem[{Fang et~al.(2025{\natexlab{b}})Fang, Liang, Wang, Wu, Qiao, Xie, Huang, Chen, and Zhang}]{fang2025memp}
Runnan Fang, Yuan Liang, Xiaobin Wang, Jialong Wu, Shuofei Qiao, Pengjun Xie, Fei Huang, Huajun Chen, and Ningyu Zhang. 2025{\natexlab{b}}.
\newblock Memp: Exploring agent procedural memory.
\newblock \emph{arXiv preprint arXiv:2508.06433}.

\bibitem[{Gao et~al.(2025)Gao, Geng, Hua, Hu, Juan, Liu, Liu, Qiu, Qi, Wu et~al.}]{gao2025survey}
Huan-ang Gao, Jiayi Geng, Wenyue Hua, Mengkang Hu, Xinzhe Juan, Hongzhang Liu, Shilong Liu, Jiahao Qiu, Xuan Qi, Yiran Wu, et~al. 2025.
\newblock A survey of self-evolving agents: On path to artificial super intelligence.
\newblock \emph{arXiv preprint arXiv:2507.21046}.

\bibitem[{Ji-An et~al.(2025)Ji-An, Mattar, Xiong, Benna, and Wilson}]{ji2025MetaCognitiveMonitoring}
Li~Ji-An, Marcelo~G Mattar, Hua-Dong Xiong, Marcus~K Benna, and Robert~C Wilson. 2025.
\newblock Language models are capable of metacognitive monitoring and control of their internal activations.
\newblock \emph{ArXiv}, pages arXiv--2505.

\bibitem[{Jimenez et~al.(2023)Jimenez, Yang, Wettig, Yao, Pei, Press, and Narasimhan}]{jimenez2023swe}
Carlos~E Jimenez, John Yang, Alexander Wettig, Shunyu Yao, Kexin Pei, Ofir Press, and Karthik Narasimhan. 2023.
\newblock Swe-bench: Can language models resolve real-world github issues?
\newblock \emph{arXiv preprint arXiv:2310.06770}.

\bibitem[{Karpukhin et~al.(2020)Karpukhin, Oguz, Min, Lewis, Wu, Edunov, Chen, and Yih}]{karpukhin2020dense}
Vladimir Karpukhin, Barlas Oguz, Sewon Min, Patrick Lewis, Ledell Wu, Sergey Edunov, Danqi Chen, and Wen-tau Yih. 2020.
\newblock Dense passage retrieval for open-domain question answering.
\newblock In \emph{Proceedings of the 2020 conference on empirical methods in natural language processing (EMNLP)}, pages 6769--6781.

\bibitem[{Kazemi et~al.(2025)Kazemi, Fatemi, Bansal, Palowitch, Anastasiou, Mehta, Jain, Aglietti, Jindal, Chen et~al.}]{kazemi2025big}
Mehran Kazemi, Bahare Fatemi, Hritik Bansal, John Palowitch, Chrysovalantis Anastasiou, Sanket~Vaibhav Mehta, Lalit~K Jain, Virginia Aglietti, Disha Jindal, Yuanzhu~Peter Chen, et~al. 2025.
\newblock Big-bench extra hard.
\newblock In \emph{Proceedings of the 63rd Annual Meeting of the Association for Computational Linguistics (Volume 1: Long Papers)}, pages 26473--26501.

\bibitem[{K{\i}c{\i}man et~al.(2023)K{\i}c{\i}man, Ness, Sharma, and Tan}]{kiciman2023causal}
Emre K{\i}c{\i}man, Robert Ness, Amit Sharma, and Chenhao Tan. 2023.
\newblock Causal reasoning and large language models: Opening a new frontier for causality.
\newblock \emph{arXiv preprint arXiv:2305.00050}.

\bibitem[{Li et~al.(2025)Li, Li, Dong, Zhang, Zhang, Liu, Wang, Tang, and Liu}]{li2025AdaptiveToolUse}
Wenjun Li, Dexun Li, Kuicai Dong, Cong Zhang, Hao Zhang, Weiwen Liu, Yasheng Wang, Ruiming Tang, and Yong Liu. 2025.
\newblock Adaptive tool use in large language models with meta-cognition trigger.
\newblock \emph{arXiv preprint arXiv:2502.12961}.

\bibitem[{Liu et~al.(2025)Liu, Mei, Lin, Xue, Wang, Xu, Wu, Zhang, Lin, Dong et~al.}]{liu2025deepseek}
Aixin Liu, Aoxue Mei, Bangcai Lin, Bing Xue, Bingxuan Wang, Bingzheng Xu, Bochao Wu, Bowei Zhang, Chaofan Lin, Chen Dong, et~al. 2025.
\newblock Deepseek-v3. 2: Pushing the frontier of open large language models.
\newblock \emph{arXiv preprint arXiv:2512.02556}.

\bibitem[{Liu et~al.(2023)Liu, Yu, Gao, Xie, Liao, Wu, and Wang}]{liu2023llm}
Jijia Liu, Chao Yu, Jiaxuan Gao, Yuqing Xie, Qingmin Liao, Yi~Wu, and Yu~Wang. 2023.
\newblock Llm-powered hierarchical language agent for real-time human-ai coordination.
\newblock \emph{arXiv preprint arXiv:2312.15224}.

\bibitem[{Liu et~al.(2026)Liu, Zhou, Xu, Huang, Wang, Zhang, Poon, and Chen}]{liu2026metascale}
Qin Liu, Wenxuan Zhou, Nan Xu, James~Y Huang, Fei Wang, Sheng Zhang, Hoifung Poon, and Muhao Chen. 2026.
\newblock Metascale: Test-time scaling with evolving meta-thoughts.
\newblock In \emph{Findings of the Association for Computational Linguistics: ACL 2026}, pages 11828--11842.

\bibitem[{Majumder et~al.(2023)Majumder, Mishra, Jansen, Tafjord, Tandon, Zhang, Callison-Burch, and Clark}]{majumder2023clin}
Bodhisattwa~Prasad Majumder, Bhavana~Dalvi Mishra, Peter Jansen, Oyvind Tafjord, Niket Tandon, Li~Zhang, Chris Callison-Burch, and Peter Clark. 2023.
\newblock Clin: A continually learning language agent for rapid task adaptation and generalization.
\newblock \emph{arXiv preprint arXiv:2310.10134}.

\bibitem[{Nie et~al.(2024)Nie, Zhang, Amdekar, Piech, Hashimoto, and Gerstenberg}]{nie2024moca}
Allen Nie, Yuhui Zhang, Atharva~Shailesh Amdekar, Chris Piech, Tatsunori~B Hashimoto, and Tobias Gerstenberg. 2024.
\newblock Moca: Measuring human-language model alignment on causal and moral judgment tasks.
\newblock \emph{Advances in Neural Information Processing Systems}, 36.

\bibitem[{OpenAI(2024)}]{web:openai:gpt4omini}
OpenAI. 2024.
\newblock \href {https://openai.com/index/gpt-4o-mini-advancing-cost-efficient-intelligence/} {Gpt-4o mini: advancing cost-efficient intelligence}.

\bibitem[{{OpenAI}(2024)}]{openai2024embedding}
{OpenAI}. 2024.
\newblock \href {https://openai.com/index/new-embedding-models-and-api-updates/} {New embedding models and api updates}.
\newblock January 25, 2024.

\bibitem[{OpenAI(2025)}]{web:openai:gpt52}
OpenAI. 2025.
\newblock \href {https://openai.com/index/gpt-5-system-card-update-gpt-5-2/} {Update to gpt-5 system card: Gpt-5.2}.

\bibitem[{Ouyang et~al.(2025)Ouyang, Yan, Hsu, Chen, Jiang, Wang, Han, Le, Daruki, Tang et~al.}]{ouyang2025reasoningbank}
Siru Ouyang, Jun Yan, I~Hsu, Yanfei Chen, Ke~Jiang, Zifeng Wang, Rujun Han, Long~T Le, Samira Daruki, Xiangru Tang, et~al. 2025.
\newblock Reasoningbank: Scaling agent self-evolving with reasoning memory.
\newblock \emph{arXiv preprint arXiv:2509.25140}.

\bibitem[{Reimers and Gurevych(2019)}]{reimers2019sentence}
Nils Reimers and Iryna Gurevych. 2019.
\newblock Sentence-bert: Sentence embeddings using siamese bert-networks.
\newblock In \emph{Proceedings of the 2019 conference on empirical methods in natural language processing and the 9th international joint conference on natural language processing (EMNLP-IJCNLP)}, pages 3982--3992.

\bibitem[{Russac et~al.(2019)Russac, Vernade, and Capp{\'e}}]{russac2019weighted}
Yoan Russac, Claire Vernade, and Olivier Capp{\'e}. 2019.
\newblock Weighted linear bandits for non-stationary environments.
\newblock \emph{Advances in Neural Information Processing Systems}, 32.

\bibitem[{Scoville et~al.(2026)Scoville, Chen, Bang, Winzeck, and Schwarz}]{scoville2026cognitive}
John Scoville, Shengzhuang Chen, Yejin Bang, Stefan Winzeck, and Jonathan~Richard Schwarz. 2026.
\newblock Cognitive demand steering for adaptive meta-reasoning in large language models.
\newblock \emph{arXiv preprint arXiv:2608.01319}.

\bibitem[{Shah et~al.(2024)Shah, Dikkala, Wang, and Panigrahy}]{shah2024causal}
Kulin Shah, Nishanth Dikkala, Xin Wang, and Rina Panigrahy. 2024.
\newblock Causal language modeling can elicit search and reasoning capabilities on logic puzzles.
\newblock \emph{arXiv preprint arXiv:2409.10502}.

\bibitem[{Shinn et~al.(2023)Shinn, Cassano, Gopinath, Narasimhan, and Yao}]{shinn2023reflexion}
Noah Shinn, Federico Cassano, Ashwin Gopinath, Karthik Narasimhan, and Shunyu Yao. 2023.
\newblock Reflexion: Language agents with verbal reinforcement learning.
\newblock \emph{Advances in Neural Information Processing Systems}, 36:8634--8652.

\bibitem[{Snell et~al.(2024)Snell, Lee, Xu, and Kumar}]{snell2024scaling}
Charlie Snell, Jaehoon Lee, Kelvin Xu, and Aviral Kumar. 2024.
\newblock Scaling llm test-time compute optimally can be more effective than scaling model parameters.
\newblock \emph{arXiv preprint arXiv:2408.03314}.

\bibitem[{Srivastava et~al.(2023)Srivastava, Rastogi, Rao, Shoeb, Abid, Fisch, Brown, Santoro, Gupta, Garriga-Alonso et~al.}]{srivastava2023beyond}
Aarohi Srivastava, Abhinav Rastogi, Abhishek Rao, Abu Awal~Md Shoeb, Abubakar Abid, Adam Fisch, Adam~R Brown, Adam Santoro, Aditya Gupta, Adri{\`a} Garriga-Alonso, et~al. 2023.
\newblock Beyond the imitation game: Quantifying and extrapolating the capabilities of language models.
\newblock \emph{Transactions on machine learning research}.

\bibitem[{Sui et~al.(2026)Sui, He, Cao, Han, Chen, and Hooi}]{sui2026meta}
Yuan Sui, Yufei He, Tri Cao, Sophia~Simeng Han, Yulin Chen, and Bryan Hooi. 2026.
\newblock Meta-reasoner: Dynamic guidance for optimized inference-time reasoning in large language models.
\newblock In \emph{Findings of the Association for Computational Linguistics: ACL 2026}, pages 13268--13286.

\bibitem[{Suzgun et~al.(2023)Suzgun, Scales, Sch{\"a}rli, Gehrmann, Tay, Chung, Chowdhery, Le, Chi, Zhou et~al.}]{suzgun2023challenging}
Mirac Suzgun, Nathan Scales, Nathanael Sch{\"a}rli, Sebastian Gehrmann, Yi~Tay, Hyung~Won Chung, Aakanksha Chowdhery, Quoc Le, Ed~Chi, Denny Zhou, et~al. 2023.
\newblock Challenging big-bench tasks and whether chain-of-thought can solve them.
\newblock In \emph{Findings of the Association for Computational Linguistics: ACL 2023}, pages 13003--13051.

\bibitem[{Tang et~al.(2025)Tang, Qin, Peng, Zhou, Shao, Du, Wei, Xia, Wu, Zhu et~al.}]{tang2025AgentKB}
Xiangru Tang, Tianrui Qin, Tianhao Peng, Ziyang Zhou, Daniel Shao, Tingting Du, Xinming Wei, Peng Xia, Fang Wu, He~Zhu, et~al. 2025.
\newblock Agent kb: Leveraging cross-domain experience for agentic problem solving.
\newblock \emph{arXiv preprint arXiv:2507.06229}.

\bibitem[{Team et~al.(2026)Team, Bai, Bai, Bao, Cai, Cao, Charles, Che, Chen, Chen et~al.}]{team2026kimi}
Kimi Team, Tongtong Bai, Yifan Bai, Yiping Bao, SH~Cai, Yuan Cao, Y~Charles, HS~Che, Cheng Chen, Guanduo Chen, et~al. 2026.
\newblock Kimi k2. 5: Visual agentic intelligence.
\newblock \emph{arXiv preprint arXiv:2602.02276}.

\bibitem[{Tyen et~al.(2023)Tyen, Mansoor, Chen, Mak, and C{\u{a}}rbune}]{tyen2023llms}
Gladys Tyen, Hassan Mansoor, Peter Chen, Tony Mak, and Victor C{\u{a}}rbune. 2023.
\newblock Llms cannot find reasoning errors, but can correct them!
\newblock \emph{arXiv preprint arXiv:2311.08516}.

\bibitem[{Wan et~al.(2026)Wan, Fang, Li, Huo, Wang, Mi, Yu, and Lyu}]{wan2026inference}
Yuxuan Wan, Tianqing Fang, Zaitang Li, Yintong Huo, Wenxuan Wang, Haitao Mi, Dong Yu, and Michael~R Lyu. 2026.
\newblock Inference-time scaling of verification: Self-evolving deep research agents via test-time rubric-guided verification.
\newblock \emph{arXiv preprint arXiv:2601.15808}.

\bibitem[{Wang et~al.(2023)Wang, Xie, Jiang, Mandlekar, Xiao, Zhu, Fan, and Anandkumar}]{wang2023voyager}
Guanzhi Wang, Yuqi Xie, Yunfan Jiang, Ajay Mandlekar, Chaowei Xiao, Yuke Zhu, Linxi Fan, and Anima Anandkumar. 2023.
\newblock Voyager: An open-ended embodied agent with large language models.
\newblock \emph{arXiv preprint arXiv:2305.16291}.

\bibitem[{Wei et~al.(2025)Wei, Sachdeva, Coleman, He, Bei, Ning, Ai, Li, He, Chi et~al.}]{wei2025evo}
Tianxin Wei, Noveen Sachdeva, Benjamin Coleman, Zhankui He, Yuanchen Bei, Xuying Ning, Mengting Ai, Yunzhe Li, Jingrui He, Ed~H Chi, et~al. 2025.
\newblock Evo-memory: Benchmarking llm agent test-time learning with self-evolving memory.
\newblock \emph{arXiv preprint arXiv:2511.20857}.

\bibitem[{White et~al.(2024)White, Dooley, Roberts, Pal, Feuer, Jain, Shwartz-Ziv, Jain, Saifullah, Naidu et~al.}]{white2024livebench}
Colin White, Samuel Dooley, Manley Roberts, Arka Pal, Ben Feuer, Siddhartha Jain, Ravid Shwartz-Ziv, Neel Jain, Khalid Saifullah, Siddartha Naidu, et~al. 2024.
\newblock Livebench: A challenging, contamination-free llm benchmark.
\newblock \emph{arXiv preprint arXiv:2406.19314}.

\bibitem[{Wu and Qu(2025)}]{wu2025MetaOptimization}
Chunlong Wu and Zhibo Qu. 2025.
\newblock Reflection-enhanced meta-optimization integrating textgrad-style prompt optimization with memory-driven self-evolution.
\newblock \emph{arXiv preprint arXiv:2508.18749}.

\bibitem[{Wu et~al.(2024)Wu, Han, Ding, Weng, Liu, Yao, Yu, and Kong}]{wu2402copilot}
Zhiyong Wu, Chengcheng Han, Zichen Ding, Zhenmin Weng, Zhoumianze Liu, Shunyu Yao, Tao Yu, and Lingpeng Kong. 2024.
\newblock Os-copilot: Towards generalist computer agents with self-improvement. corr, abs/2402.07456, 2024. doi: 10.48550.
\newblock \emph{arXiv preprint ARXIV.2402.07456}.

\bibitem[{Xu et~al.(2025)Xu, Liang, Mei, Gao, Tan, and Zhang}]{xu2025AgenticMem}
Wujiang Xu, Zujie Liang, Kai Mei, Hang Gao, Juntao Tan, and Yongfeng Zhang. 2025.
\newblock A-mem: Agentic memory for llm agents.
\newblock \emph{arXiv preprint arXiv:2502.12110}.

\bibitem[{Yang et~al.(2025)Yang, Yang, Wen, Fu, Mei, Wu, Cai, Shen, Deng, Shi et~al.}]{yang2025MUSE}
Cheng Yang, Xuemeng Yang, Licheng Wen, Daocheng Fu, Jianbiao Mei, Rong Wu, Pinlong Cai, Yufan Shen, Nianchen Deng, Botian Shi, et~al. 2025.
\newblock Learning on the job: An experience-driven self-evolving agent for long-horizon tasks.
\newblock \emph{arXiv preprint arXiv:2510.08002}.

\bibitem[{Zhang et~al.(2026{\natexlab{a}})Zhang, Zhao, Yang, Foerster, Clune, Jiang, Devlin, and Shavrina}]{zhang2026hyperagents}
Jenny Zhang, Bingchen Zhao, Wannan Yang, Jakob Foerster, Jeff Clune, Minqi Jiang, Sam Devlin, and Tatiana Shavrina. 2026{\natexlab{a}}.
\newblock Hyperagents.
\newblock \emph{arXiv preprint arXiv:2603.19461}.

\bibitem[{Zhang et~al.(2025{\natexlab{a}})Zhang, Hu, Upasani, Ma, Hong, Kamanuru, Rainton, Wu, Ji, Li et~al.}]{zhang2025ACE}
Qizheng Zhang, Changran Hu, Shubhangi Upasani, Boyuan Ma, Fenglu Hong, Vamsidhar Kamanuru, Jay Rainton, Chen Wu, Mengmeng Ji, Hanchen Li, et~al. 2025{\natexlab{a}}.
\newblock Agentic context engineering: Evolving contexts for self-improving language models.
\newblock \emph{arXiv preprint arXiv:2510.04618}.

\bibitem[{Zhang et~al.(2025{\natexlab{b}})Zhang, Wang, Zhang, Li, Song, Li, Qiu, Cao, Cai, Yao et~al.}]{zhang2025DPTAgent}
Shao Zhang, Xihuai Wang, Wenhao Zhang, Chaoran Li, Junru Song, Tingyu Li, Lin Qiu, Xuezhi Cao, Xunliang Cai, Wen Yao, et~al. 2025{\natexlab{b}}.
\newblock Leveraging dual process theory in language agent framework for real-time simultaneous human-ai collaboration.
\newblock \emph{arXiv preprint arXiv:2502.11882}.

\bibitem[{Zhang et~al.(2026{\natexlab{b}})Zhang, Wang, Zhou, Liao, Feng, Zhang, Wen, Li, Xiong, Qi et~al.}]{zhang2026memrl}
Shengtao Zhang, Jiaqian Wang, Ruiwen Zhou, Junwei Liao, Yuchen Feng, Weinan Zhang, Ying Wen, Zhiyu Li, Feiyu Xiong, Yutao Qi, et~al. 2026{\natexlab{b}}.
\newblock Memrl: Self-evolving agents via runtime reinforcement learning on episodic memory.
\newblock \emph{arXiv preprint arXiv:2601.03192}.

\bibitem[{Zhao et~al.(2024)Zhao, Huang, Xu, Lin, Liu, and Huang}]{zhao2024expel}
Andrew Zhao, Daniel Huang, Quentin Xu, Matthieu Lin, Yong-Jin Liu, and Gao Huang. 2024.
\newblock Expel: Llm agents are experiential learners.
\newblock In \emph{Proceedings of the AAAI Conference on Artificial Intelligence}, volume~38, pages 19632--19642.

\bibitem[{Zhuo et~al.(2024)Zhuo, Vu, Chim, Hu, Yu, Widyasari, Yusuf, Zhan, He, Paul et~al.}]{zhuo2024bigcodebench}
Terry~Yue Zhuo, Minh~Chien Vu, Jenny Chim, Han Hu, Wenhao Yu, Ratnadira Widyasari, Imam Nur~Bani Yusuf, Haolan Zhan, Junda He, Indraneil Paul, et~al. 2024.
\newblock Bigcodebench: Benchmarking code generation with diverse function calls and complex instructions.
\newblock \emph{arXiv preprint arXiv:2406.15877}.

\end{thebibliography}

\clearpage
\appendix
\section*{Appendix}
\section{Detailed Representation of the Meta-State.}
\label{app:meta_state}

For each decision step $t$, the Meta-Controller receives a compact
nine-dimensional representation of the retrieved experience rather than the
raw memory text:
\begin{equation}
x_t=(x_{t,1},\ldots,x_{t,8},1)\in\mathbb{R}^{9},
\end{equation}
where the final coordinate is a constant bias term. In all experiments, the
retrieval count is fixed to $M=3$ for each memory source.

Let $\mathcal{M}^{\mathrm{retr}}_{\mathrm{epi},t}$ and
$\mathcal{M}^{\mathrm{retr}}_{\mathrm{guide},t}$ denote the retrieved episodic
records and guidance entries at step $t$, sorted by retrieval
score in descending order. For an item $j$ from source
$z\in\{\mathrm{epi},\mathrm{guide}\}$, let $s^z_{t,j}$ denote its retrieval
score and let $v^z_{t,j}$ denote its quality score. The retrieval scores are
cosine similarities between the current query and the corresponding memory
entry representations computed by dense text embeddings.

For episodic memory, the quality score is mapped from the observed outcome as
\begin{equation}
v^{\mathrm{epi}}_{t,j} =
\begin{cases}
1.0, & \text{success},\\
0.5, & \text{partial or unknown outcome},\\
0.0, & \text{failure}.
\end{cases}
\end{equation}
For guidance memory, $v^{\mathrm{guide}}_{t,j}$ is computed from
available success-rate and confidence statistics. When these statistics are
unavailable, a memory-type prior is used.

Let $H_M(\cdot)$ denote the top-$M$ items in a retrieved set. When fewer than
$M$ items are available, $H_M(\cdot)$ contains all available items. The eight
state features are defined as
\begin{equation}
\begin{aligned}
x_{t,1} &=
\frac{1}{|H_M(\mathcal{M}^{\mathrm{retr}}_{\mathrm{epi},t})|}
\sum_{j\in H_M(\mathcal{M}^{\mathrm{retr}}_{\mathrm{epi},t})}
s^{\mathrm{epi}}_{t,j}, \\
x_{t,2} &=
\frac{1}{|H_M(\mathcal{M}^{\mathrm{retr}}_{\mathrm{guide},t})|}
\sum_{j\in H_M(\mathcal{M}^{\mathrm{retr}}_{\mathrm{guide},t})}
s^{\mathrm{guide}}_{t,j},\\
x_{t,3} &=
\frac{1}{|H_M(\mathcal{M}^{\mathrm{retr}}_{\mathrm{epi},t})|}
\sum_{j\in H_M(\mathcal{M}^{\mathrm{retr}}_{\mathrm{epi},t})}
v^{\mathrm{epi}}_{t,j}, \\
x_{t,4} &=
\frac{1}{|\mathcal{M}^{\mathrm{retr}}_{\mathrm{epi},t}|}
\sum_{j\in\mathcal{M}^{\mathrm{retr}}_{\mathrm{epi},t}}
s^{\mathrm{epi}}_{t,j}v^{\mathrm{epi}}_{t,j},\\
x_{t,5} &=
\frac{1}{|H_M(\mathcal{M}^{\mathrm{retr}}_{\mathrm{guide},t})|}
\sum_{j\in H_M(\mathcal{M}^{\mathrm{retr}}_{\mathrm{guide},t})}
v^{\mathrm{guide}}_{t,j}, \\
x_{t,6} &=
\frac{1}{|\mathcal{M}^{\mathrm{retr}}_{\mathrm{guide},t}|}
\sum_{j\in\mathcal{M}^{\mathrm{retr}}_{\mathrm{guide},t}}
s^{\mathrm{guide}}_{t,j}v^{\mathrm{guide}}_{t,j},\\
x_{t,7} &=
\frac{2\min(n_{t,s},n_{t,f}+\rho n_{t,p})}
{n_{t,s}+n_{t,f}+\rho n_{t,p}}, \\
x_{t,8} &=
\frac{-\sum_o p_{t,o}\log p_{t,o}}
{\log|\{o:p_{t,o}>0\}|}.
\end{aligned}
\end{equation}

Here, $n_{t,s}$, $n_{t,f}$, and $n_{t,p}$ denote the numbers of successful,
failed, and partial episodic outcomes among the retrieved records. Partial
outcomes contribute to the failure side with weight $\rho=0.5$. In particular,
we define
\begin{equation}
\tilde{n}_{t,f}=n_{t,f}+\rho n_{t,p}.
\end{equation}
Thus, $x_{t,7}$ measures the balance between successful episodic evidence and
failure-side evidence, where partial outcomes contribute with half the weight
of failures.

For the entropy feature, $p_{t,o}$ denotes the empirical proportion of outcome
type $o$ among the retrieved episodic records:
\begin{equation}
\begin{aligned}
p_{t,o}
&=
\frac{n_{t,o}}
{\sum_{o'\in\mathcal{O}}n_{t,o'}}, \\
\mathcal{O}
&=
\{\text{success},\text{failure},\text{partial}\}.
\end{aligned}
\end{equation}
If no episodic record is retrieved, we set $x_{t,7}=x_{t,8}=0$. If the
retrieved episodic records contain only one nonzero outcome type, we set
$x_{t,8}=0$. Otherwise, $x_{t,8}$ is the normalized entropy of the retrieved
outcome types.

If a memory source is empty, its corresponding source-level mean features and
relevance-weighted features are set to zero. Therefore, when episodic memory is
empty, $x_{t,1}$, $x_{t,3}$, and $x_{t,4}$ are set to zero. When
guidance memory is empty, $x_{t,2}$, $x_{t,5}$, and $x_{t,6}$ are set to zero.

The first eight coordinates are normalized online using running z-score
statistics with symmetric clipping:
\begin{equation}
\begin{aligned}
\tilde{x}_{t,i}
&=
\operatorname{clip}
\left(
\frac{x_{t,i}-\mu_i}
{\max(\sigma_i,\epsilon)},
-c,c
\right),
\qquad
i=1,\ldots,8, \\
\tilde{x}_{t,9}&=1.  
\end{aligned}
\end{equation}
The bias coordinate is not normalized. The features $x_{t,7}$ and $x_{t,8}$
are exposed only to the DISTILL action $a_{\textsc{DISTILL}}$ through an
action-specific feature mask.

The state therefore summarizes \emph{Relevance} through retrieval scores,
\emph{Quality} through outcome- and confidence-based scores and their
relevance-weighted counterparts, and \emph{Diversity} through the balance and
normalized entropy of episodic outcome types. Here, Diversity refers specifically
to episodic outcome diversity, not semantic diversity among retrieved entries.
The controller uses this compact summary to select among
DIRECT, EPISODE, GUIDE, and
DISTILL without directly accessing the retrieved memory text.

\section{Action Selection Strategies: Thompson Sampling and LinUCB}
\label{app:action_selection}

For the contextual-bandit action-selection algorithms, we maintain an action-wise linear reward model. At each round $t$, the Meta-Controller observes a meta-state $x_t\in\mathbb{R}^d$ and selects an action $a_t\in\mathcal{A}$. Let $r_t\in[0,1]$ denote the immediate reward.

For each action $a\in\mathcal{A}$, the sufficient statistics are initialized as $A_a^{(1)}=\lambda I_d$ and $b_a^{(1)}=0$. After observing the reward, the statistics are updated with a forgetting factor $\gamma \in (0, 1]$:
\begin{equation}
A_a^{(t+1)} = \gamma A_a^{(t)} + x_t x_t^\top, \qquad b_a^{(t+1)} = \gamma b_a^{(t)} + r_t x_t.
\end{equation}
The posterior mean is given by $\hat{\theta}_a^{(t)} = \bigl(A_a^{(t)}\bigr)^{-1}b_a^{(t)}$, and the posterior covariance proxy is $\Sigma_a^{(t)} = v^2\bigl(A_a^{(t)}\bigr)^{-1}$, where $v>0$ is a scale parameter.

\paragraph{Thompson Sampling (\method-TS).}
Thompson Sampling balances exploration and exploitation by sampling from the posterior distribution. At round $t$, for each action $a\in\mathcal{A}$, we draw a parameter sample:
\begin{equation}
\label{eq:ThompsonSampling_impl}
\tilde{\theta}_a^{(t)} \sim \mathcal{N}\!\left(\hat{\theta}_a^{(t)},\,\Sigma_a^{(t)}\right).
\end{equation}

The action is then selected by maximizing the sampled score:
\begin{equation}
\label{eq:decide_action}
a_t = \arg\max_{a\in\mathcal{A}} x_t^\top \tilde{\theta}_a^{(t)}.
\end{equation}

\paragraph{LinUCB (\method-UCB).}
The optimistic LinUCB-style rule selects the action that maximizes an upper confidence bound on the expected reward. At round $t$, the action is selected as:
\begin{equation}
a_t = \arg\max_{a\in\mathcal{A}} \left\{ x_t^\top \hat{\theta}_a^{(t)} + \alpha \sqrt{x_t^\top \bigl(A_a^{(t)}\bigr)^{-1} x_t} \right\},
\end{equation}
where $\alpha > 0$ is an exploration hyperparameter that controls the confidence interval width.

\section{Heuristic Baseline Details}
\label{app:heuristic_baseline}

The \textit{heuristic} baseline employs a fixed, rule-based policy to select actions based solely on retrieval scores, without any online learning or parameter updates. The decision mechanism evaluates the available signals against predefined thresholds to compute a heuristic score for each possible action, ultimately selecting the action with the maximum score.

Let $s_{\mathrm{epi}}$ and $s_{\mathrm{guide}}$ denote the mean retrieval scores for the episodic records and guidance entries, respectively. The heuristic relies on three configurable hyperparameters: the weak threshold $\tau_{w}$, the use threshold $\tau_{u}$, and the margin threshold $\tau_{m}$. To ensure numerical stability, we define a small constant $\epsilon = 10^{-8}$.

The heuristic scores for the four actions in the action space $\mathcal{A}$ are computed as follows:

\paragraph{DIRECT ($a_{\textsc{DIRECT}}$).}
The score for DIRECT is inversely proportional to the maximum available retrieval score, encouraging the agent to skip when all retrieval signals are weak:
\begin{equation}
    \text{score}_{\textsc{DIRECT}} = \max\left(0, 1 - \frac{\max(s_{\mathrm{epi}}, s_{\mathrm{guide}})}{\max(\tau_w, \epsilon)}\right)
\end{equation}

\paragraph{EPISODE ($a_{\textsc{EPISODE}}$).}
The score for utilizing an episodic record depends on its absolute signal strength and its relative advantage over the guidance entry signal. We define the episodic strength $\text{str}_{\mathrm{epi}}$ and episodic advantage $\text{adv}_{\mathrm{epi}}$ as:
\begin{align}
    \text{str}_{\mathrm{epi}} &= \frac{\max(0, s_{\mathrm{epi}} - \tau_w)}{\max(\tau_u - \tau_w, \epsilon)} \\
    \text{adv}_{\mathrm{epi}} &= \max(0, s_{\mathrm{epi}} - s_{\mathrm{guide}})
\end{align}
The final score is a weighted combination that rewards both absolute strength and relative advantage:
\begin{equation}
\begin{aligned}
\text{score}_{\textsc{EPISODE}}
&= \min(\text{str}_{\mathrm{epi}}, 1)\\
& \cdot \left(
0.7 + 0.3 \cdot
\min\left(
\frac{\text{adv}_{\mathrm{epi}}}
{\max(\tau_m, \epsilon)}, 1
\right)
\right).
\end{aligned}
\end{equation}

\paragraph{GUIDE ($a_{\textsc{GUIDE}}$).}
Symmetrically, the score for using a guidance entry is computed based on its strength $\text{str}_{\mathrm{guide}}$ and advantage $\text{adv}_{\mathrm{guide}}$:
\begin{align}
    \text{str}_{\mathrm{guide}} &= \frac{\max(0, s_{\mathrm{guide}} - \tau_w)}{\max(\tau_u - \tau_w, \epsilon)} \\
    \text{adv}_{\mathrm{guide}} &= \max(0, s_{\mathrm{guide}} - s_{\mathrm{epi}})
\end{align}
\begin{equation}
\begin{aligned}
\text{score}_{\textsc{GUIDE}}
&= \min(\text{str}_{\mathrm{guide}}, 1)\\
& \cdot \left(
0.7 + 0.3 \cdot
\min\left(
\frac{\text{adv}_{\mathrm{guide}}}
{\max(\tau_m, \epsilon)}, 1
\right)
\right).
\end{aligned}
\end{equation}

\paragraph{DISTILL ($a_{\textsc{DISTILL}}$).}
The score for distilling a new guidance entry is determined by a piecewise function based on the best available retrieval score, $s_{\max} = \max(s_{\mathrm{epi}}, s_{\mathrm{guide}})$. Distillation is favored when the retrieval scores are moderate (between $\tau_w$ and $\tau_u$), and penalized if the scores are either too weak (indicating poor context) or sufficiently strong (indicating distillation is unnecessary):
\begin{equation}
\begin{aligned}
&\text{score}_{\textsc{DISTILL}}
=\\
&\begin{cases}
0, & s_{\max}<\tau_w,\\
\max(0,\,0.5-(s_{\max}-\tau_u)),
    & s_{\max}\ge\tau_u,\\
\dfrac{s_{\max}-\tau_w}
{\max(\tau_u-\tau_w,\epsilon)},
     \text{otherwise}.
\end{cases}
\end{aligned}
\end{equation}

\paragraph{Decision Rule.}
The final action $a^*$ is selected greedily according to the highest computed heuristic score:
\begin{equation}
    a^* = \arg\max_{a \in \mathcal{A}} \text{score}_a
\end{equation}

\section{Full Routing Ablation Across Backbones}
\label{app:routing_ablation}

\Cref{tab:ablation} in the main text reports the routing ablation on the reference backbone, GPT-5.2; the table below gives the same comparison for all four backbones, which is what the cross-backbone counts in \Cref{sec:ablation} are computed from. Two readings are worth noting. First, the ordering of the two fixed policies is not stable: on BBEH, Fixed-$a_{\textsc{EPISODE}}$ leads Fixed-$a_{\textsc{GUIDE}\&\textsc{DISTILL}}$ by 7.2 points with GPT-5.2, while the order reverses with DeepSeek-V3.2, where Fixed-$a_{\textsc{GUIDE}\&\textsc{DISTILL}}$ leads by 4.7 points. Second, no single fixed policy leads on more than three of the four backbones of any benchmark, so the reversal is not an artifact of one backbone LLM. Across the same grid, the two learned variants split the benchmark--backbone pairs almost evenly: \method-TS leads on 8 of the 16, \method-UCB on 7, with one tie.

\begin{table*}[htbp]
\vspace{-8pt}
\centering
\scriptsize
\setlength{\tabcolsep}{4pt}
\renewcommand{\arraystretch}{1.15}
\resizebox{\textwidth}{!}{%
\begin{tabular}{>{\centering\arraybackslash}m{1.6cm}
                >{\centering\arraybackslash}m{2.0cm}
                >{\centering\arraybackslash}m{1.76cm}
                >{\centering\arraybackslash}m{1.96cm}
                >{\centering\arraybackslash}m{1.76cm}
                >{\centering\arraybackslash}m{1.78cm}
                >{\centering\arraybackslash}m{1.78cm}}
\toprule
\multirow{2}{*}{\centering\textbf{Benchmark}} & \multirow{2}{*}{\centering\textbf{Model}} & \multicolumn{5}{c}{\textbf{OSR $\uparrow$ / CSR $\uparrow$}} \\
\cline{3-7}
 & & \textbf{Fixed-}$a_{\textsc{EPISODE}}$
     & \textbf{Fixed-}$a_{\textsc{GUIDE}\&\textsc{DISTILL}}$
     & \method-HEU
     & \method-TS
     & \method-UCB \\
\midrule

\multirow{4}{*}{\centering\makecell{\textbf{Multi-}\\ \textbf{Challenge}}}
& GPT-5.2          & 56.0\% / 91.2\% & 63.7\% / 94.5\% & 64.5\% / 95.2\% & \textbf{67.4\%} / \textbf{97.1\%} & 66.3\% / 95.6\% \\
& Deepseek-V3.2    & 68.9\% / \textbf{97.4\%} & 70.3\% / \textbf{97.4\%} & 69.2\% / \textbf{97.4\%} & \textbf{70.7\%} / 95.6\% & 67.4\% / 96.7\% \\
& Deepseek-V4-pro  & 65.6\% / 91.9\% & 67.0\% / \textbf{95.2\%} & 66.3\% / 93.8\% & \textbf{69.6\%} / 93.8\% & 64.8\% / 91.2\% \\
& Kimi-K2.5        & \textbf{73.3\%} / \textbf{96.7\%} & 72.2\% / \textbf{96.7\%} & 72.2\% / \textbf{96.7\%} & 72.2\% / 95.6\% & \textbf{73.3\%} / \textbf{96.7\%} \\
\midrule

\multirow{4}{*}{\centering\textbf{BBEH}}
& GPT-5.2          & 50.3\% / 76.8\% & 43.1\% / 73.8\% & 45.6\% / 71.7\% & \textbf{53.3\%} / \textbf{78.7\%} & 46.8\% / 71.0\% \\
& Deepseek-V3.2    & 42.5\% / 71.3\% & \textbf{47.2\%} / 73.1\% & 46.1\% / 71.4\% & 45.8\% / \textbf{73.8\%} & 46.3\% / 73.3\% \\
& Deepseek-V4-pro  & 47.4\% / 79.8\% & 46.4\% / \textbf{80.0\%} & \textbf{47.5\%} / 79.0\% & 47.3\% / 79.8\% & 47.4\% / \textbf{80.0\%} \\
& Kimi-K2.5        & 47.9\% / 80.3\% & 46.8\% / 77.8\% & 48.3\% / 77.6\% & \textbf{48.7\%} / 80.7\% & 48.4\% / \textbf{81.2\%} \\
\midrule

\multirow{4}{*}{\centering\makecell{\textbf{BigCode-}\\ \textbf{Bench}}}
& GPT-5.2          & 49.0\% / 59.0\% & 49.6\% / 58.0\% & 47.4\% / 57.4\% & 49.6\% / 58.9\% & \textbf{51.3\%} / \textbf{59.2\%} \\
& Deepseek-V3.2    & 46.9\% / 65.2\% & 47.0\% / 63.6\% & 47.6\% / 65.2\% & \textbf{47.7\%} / 65.2\% & 47.1\% / \textbf{65.5\%} \\
& Deepseek-V4-pro  & \textbf{45.7\%} / 72.2\% & 43.8\% / 70.4\% & 44.0\% / \textbf{72.5\%} & 43.3\% / 70.7\% & 44.1\% / 69.7\% \\
& Kimi-K2.5        & 50.0\% / 71.0\% & 49.5\% / 69.1\% & \textbf{50.6\%} / \textbf{71.3\%} & 48.5\% / 68.3\% & 49.1\% / 69.7\% \\
\midrule

\multirow{4}{*}{\centering\makecell{\textbf{SWE-bench} \\ \textbf{Verified}}}
& GPT-5.2          & 64.2\% / \textbf{64.8\%} & 64.4\% / \textbf{64.8\%} & 64.0\% / \textbf{64.8\%} & \textbf{64.8\%} / \textbf{64.8\%} & \textbf{64.8\%} / \textbf{64.8\%} \\
& Deepseek-V3.2    & 52.2\% / 53.2\% & \textbf{53.0\%} / 53.8\% & 51.4\% / 53.4\% & 52.4\% / 53.8\% & \textbf{53.0\%} / \textbf{54.0\%} \\
& Deepseek-V4-pro  & 58.4\% / 60.8\% & 58.6\% / \textbf{61.2\%} & 57.0\% / 59.6\% & \textbf{58.8\%} / 60.4\% & 58.4\% / \textbf{61.2\%} \\
& Kimi-K2.5        & 48.6\% / 48.8\% & 48.6\% / \textbf{50.6\%} & 48.0\% / 48.6\% & \textbf{48.8\%} / 49.0\% & 48.6\% / 48.6\% \\
\bottomrule
\end{tabular}
}
\caption{Full routing ablation of \method on four backbones, OSR / CSR (\%). Fixed-$a_{\textsc{EPISODE}}$ and Fixed-$a_{\textsc{GUIDE}\&\textsc{DISTILL}}$ always commit to one form of experience; \method-HEU selects actions with hand-crafted rules. The GPT-5.2 block is reproduced from \Cref{tab:ablation} for reference.}
\label{tab:routing_ablation_full}
\vspace{-12pt}
\end{table*}

\section{Additional Theoretical Result for the Meta-Controller}
\label{app:theorem}

For the theoretical analysis, we consider the stationary regime and set the forgetting factor to $\gamma=1$.
Let $\mathcal{A}$ denote the set of cognitive actions, with $K:=|\mathcal{A}|=4$.
At each round $t$, the Meta-Controller observes a meta-state $x_t\in\mathbb{R}^d$ and selects an action $a_t\in\mathcal{A}$.
The reward $r_t\in[0,1]$ denotes the immediate utility of the current query after executing $a_t$.

For each action $a\in\mathcal{A}$, initialize
\begin{equation}
A_a^{(1)}=\lambda I_d,\qquad b_a^{(1)}=0,
\end{equation}
and define
\begin{equation}
\begin{aligned}
A_a^{(t)}
=
\lambda I_d+\sum_{s=1}^{t-1}\mathbf{1}\{a_s=a\}x_sx_s^\top, \\
b_a^{(t)}
=
\sum_{s=1}^{t-1}\mathbf{1}\{a_s=a\}r_sx_s,
\end{aligned}
\end{equation}
\begin{equation}
\begin{aligned}
\hat{\theta}_a^{(t)}
=
\bigl(A_a^{(t)}\bigr)^{-1}b_a^{(t)}, \\
N_a(t-1)
=
\sum_{s=1}^{t-1}\mathbf{1}\{a_s=a\}.
\end{aligned}
\end{equation}

In the main text, $\mu_a(x_t)$ is used as a shorthand for the expected immediate reward of action $a$ under the current meta-state. 
For the theoretical analysis, we make this quantity explicit by assuming that for each action $a\in\mathcal{A}$, there exists an unknown parameter $\theta_a^\star\in\mathbb{R}^d$ such that
\begin{equation}
\begin{aligned}
\mu_a(x_t)
:=
\mathbb{E}[r_t\mid \mathcal{F}_{t-1},x_t,a_t=a]
=
x_t^\top\theta_a^\star, \\
\|\theta_a^\star\|_2\le S,
\end{aligned}
\end{equation}
and that $\|x_t\|_2\le L$ for all $t$.

Moreover, letting
\begin{equation}
\kappa_t:=r_t-x_t^\top\theta_{a_t}^\star,
\end{equation}
we assume that $\kappa_t$ is conditionally $R$-sub-Gaussian, i.e.,
\begin{equation}
\begin{aligned}
\mathbb{E}\!\left[\exp(\alpha \kappa_t)\mid \mathcal{F}_{t-1},x_t,a_t\right]
\le
\exp\!\left(\frac{\alpha^2R^2}{2}\right), \\
\forall \alpha\in\mathbb{R}.
\end{aligned}
\end{equation}

The main paper implements the Meta-Controller via Thompson Sampling on the same action-wise linear reward model.
Concretely, for each action $a\in\mathcal{A}$, define the posterior mean
\begin{equation}
\hat{\theta}_a^{(t)}=\bigl(A_a^{(t)}\bigr)^{-1}b_a^{(t)},
\end{equation}
and the posterior covariance proxy
\begin{equation}
\Sigma_a^{(t)}=v^2\bigl(A_a^{(t)}\bigr)^{-1},
\end{equation}
where $v>0$ is a scale parameter.
At round $t$, Thompson Sampling draws
\begin{equation}
\label{eq:ThompsonSampling}
\tilde{\theta}_a^{(t)} \sim \mathcal{N}\!\left(\hat{\theta}_a^{(t)},\,\Sigma_a^{(t)}\right),
\qquad \forall a\in\mathcal{A},
\end{equation}
computes the sampled score
\begin{equation}
\tilde{r}_a^{(t)} = x_t^\top \tilde{\theta}_a^{(t)},
\end{equation}
and selects the action
\begin{equation}
a_t
=
\arg\max_{a\in\mathcal{A}} \tilde{r}_a^{(t)}.
\end{equation}
After observing the reward $r_t$, the sufficient statistics of the selected action are updated according to the equations above.

For the regret analysis below, we consider an optimistic instantiation built on the same action-wise linear estimator, which yields a shorter and more standard finite-time proof.

\begin{lemma}[Uniform confidence ellipsoid for the action-wise linear reward models]
\label{lem:confidence_ellipsoid}
For any $\delta\in(0,1)$, define
\begin{equation}
\begin{aligned}
\beta_a^{(t)}(\delta)
&:=
R\sqrt{
d\log\!\left(1+\frac{N_a(t-1)L^2}{\lambda d}\right)
+
2\log\!\frac{K}{\delta}
} \\
&+
\sqrt{\lambda}\,S. 
\end{aligned}
\end{equation}
Then, with probability at least $1-\delta$, the following event holds simultaneously for all $t\ge 1$ and all $a\in\mathcal{A}$:
\begin{equation}
\bigl\|\hat{\theta}_a^{(t)}-\theta_a^\star\bigr\|_{A_a^{(t)}}
\le
\beta_a^{(t)}(\delta).
\end{equation}
\end{lemma}

\begin{proof}
Fix an action $a\in\mathcal{A}$.
Let $\tau_{a,1}<\tau_{a,2}<\cdots$ denote the rounds at which action $a$ is selected, and let $n:=N_a(t-1)$.
Define
\begin{equation}
\begin{aligned}
X_{a,i}:=x_{\tau_{a,i}},
\qquad
\xi_{a,i}:=r_{\tau_{a,i}}-X_{a,i}^\top\theta_a^\star, \\
i=1,\dots,n.
\end{aligned}
\end{equation}
Then
\begin{equation}
V_{a,n}:=\lambda I_d+\sum_{i=1}^n X_{a,i}X_{a,i}^\top
=
A_a^{(t)}.
\end{equation}
By the self-normalized concentration inequality for linear bandits \cite{abbasi2011improved}, with probability at least $1-\delta/K$, simultaneously for all $n\ge 0$,
\begin{equation}
\begin{aligned}
&\left\|
\sum_{i=1}^n \xi_{a,i}X_{a,i}
\right\|_{V_{a,n}^{-1}} \\
&\le
R\sqrt{
2\log\!\left(
\frac{\det(V_{a,n})^{1/2}}
{\det(\lambda I_d)^{1/2}}
\cdot
\frac{K}{\delta}
\right)
}.
\end{aligned}
\end{equation}
Also,
\begin{equation}
\begin{aligned}
b_a^{(t)}
&=
\sum_{i=1}^n r_{\tau_{a,i}}X_{a,i} \\
&=
\left(\sum_{i=1}^n X_{a,i}X_{a,i}^\top\right)\theta_a^\star
+
\sum_{i=1}^n \xi_{a,i}X_{a,i},
\end{aligned}
\end{equation}
hence
\begin{equation}
A_a^{(t)}\bigl(\hat{\theta}_a^{(t)}-\theta_a^\star\bigr)
=
\sum_{i=1}^n \xi_{a,i}X_{a,i}
-
\lambda\theta_a^\star.
\end{equation}
Therefore,
\begin{equation}
\begin{aligned}
&\bigl\|\hat{\theta}_a^{(t)}-\theta_a^\star\bigr\|_{A_a^{(t)}} \\
&\le
\left\|
\sum_{i=1}^n \xi_{a,i}X_{a,i}
\right\|_{(A_a^{(t)})^{-1}}
+
\bigl\|\lambda\theta_a^\star\bigr\|_{(A_a^{(t)})^{-1}}.
\end{aligned}
\end{equation}
Since $A_a^{(t)}\succeq \lambda I_d$ and $\|\theta_a^\star\|_2\le S$,
\begin{equation}
\bigl\|\lambda\theta_a^\star\bigr\|_{(A_a^{(t)})^{-1}}
\le
\sqrt{\lambda}\,S.
\end{equation}
Moreover,
\begin{equation}
\begin{aligned}
\det(A_a^{(t)})
&\le
\left(\frac{\mathrm{tr}(A_a^{(t)})}{d}\right)^d\\
&\le
\left(\lambda+\frac{N_a(t-1)L^2}{d}\right)^d,
\end{aligned}
\end{equation}
which implies
\begin{equation}
\frac{\det(A_a^{(t)})}{\det(\lambda I_d)}
\le
\left(1+\frac{N_a(t-1)L^2}{\lambda d}\right)^d.
\end{equation}
Combining the above inequalities yields
\begin{equation}
\begin{aligned}
&\bigl\|\hat{\theta}_a^{(t)}-\theta_a^\star\bigr\|_{A_a^{(t)}}\\
&\le R\Biggl[
d\log\!\left(1+\frac{N_a(t-1)L^2}{\lambda d}\right)\\
&\qquad+2\log\!\frac{K}{\delta}
\Biggr]^{1/2}
+\sqrt{\lambda}\,S.
\end{aligned}
\end{equation}
Finally, applying a union bound over all $a\in\mathcal{A}$ proves the lemma.
\end{proof}

For the regret analysis, consider the optimistic instantiation
\begin{equation}
a_t
=
\arg\max_{a\in\mathcal{A}}
\left\{
x_t^\top\hat{\theta}_a^{(t)}
+
\bar{\beta}_a^{(t)}(\delta)\,
\|x_t\|_{(A_a^{(t)})^{-1}}
\right\},
\end{equation}
where
\begin{equation}
\bar{\beta}_a^{(t)}(\delta):=\max\{1,\beta_a^{(t)}(\delta)\}.
\end{equation}
Let
\begin{equation}
a_t^\star:=\arg\max_{a\in\mathcal{A}}\mu_a(x_t),
\end{equation}
and define the cumulative pseudo-regret
\begin{equation}
R_T
:=
\sum_{t=1}^T
\Bigl(
\mu_{a_t^\star}(x_t)-\mu_{a_t}(x_t)
\Bigr).
\end{equation}

\begin{theorem}[Sublinear cumulative pseudo-regret of the optimistic Meta-Controller]
\label{thm:regret}
Suppose $\lambda\ge L^2$.
Then, with probability at least $1-\delta$,
\begin{equation}
R_T
\le
2\bar{\beta}_T(\delta)\,
\sqrt{
2KdT\log\!\left(1+\frac{TL^2}{\lambda d}\right)
},
\end{equation}
where
\begin{equation}
\begin{aligned}
\bar{\beta}_T(\delta)
&:= \max\Biggl\{1,\\
&\quad R\sqrt{
d\log\!\left(1+\frac{TL^2}{\lambda d}\right)
+2\log\!\frac{K}{\delta}
}\\
&\quad+\sqrt{\lambda}S
\Biggr\}.
\end{aligned}
\end{equation}
In particular, since $K=4$ is fixed, $R_T=\tilde{O}(d\sqrt{T})$.
\end{theorem}

\begin{proof}
Let $\mathcal{E}_\delta$ denote the high-probability event in Lemma~\ref{lem:confidence_ellipsoid}.
Fix a round $t$ and define
\begin{equation}
U_a^{(t)}(x_t)
:=
x_t^\top\hat{\theta}_a^{(t)}
+
\bar{\beta}_a^{(t)}(\delta)\,
\|x_t\|_{(A_a^{(t)})^{-1}}.
\end{equation}
By Lemma~\ref{lem:confidence_ellipsoid}, on $\mathcal{E}_\delta$, for every action $a$,
\begin{equation}
\begin{aligned}
\mu_a(x_t)
&=
x_t^\top\theta_a^\star
\le
x_t^\top\hat{\theta}_a^{(t)}
+
\beta_a^{(t)}(\delta)\,
\|x_t\|_{(A_a^{(t)})^{-1}} \\
&\le
U_a^{(t)}(x_t).
\end{aligned}
\end{equation}
Hence
\begin{equation}
\mu_{a_t^\star}(x_t)\le U_{a_t^\star}^{(t)}(x_t)\le U_{a_t}^{(t)}(x_t),
\end{equation}
which implies
\begin{equation}
\mu_{a_t^\star}(x_t)-\mu_{a_t}(x_t)
\le
U_{a_t}^{(t)}(x_t)-\mu_{a_t}(x_t).
\end{equation}
Again by Lemma~\ref{lem:confidence_ellipsoid},
\begin{equation}
\begin{aligned}
U_{a_t}^{(t)}(x_t)-\mu_{a_t}(x_t)
&\le
2\bar{\beta}_{a_t}^{(t)}(\delta)\,
\|x_t\|_{(A_{a_t}^{(t)})^{-1}} \\
&\le
2\bar{\beta}_T(\delta)\,
\|x_t\|_{(A_{a_t}^{(t)})^{-1}}.  
\end{aligned}
\end{equation}
Summing over $t=1,\dots,T$ gives
\begin{equation}
R_T
\le
2\bar{\beta}_T(\delta)
\sum_{t=1}^T
\|x_t\|_{(A_{a_t}^{(t)})^{-1}}.
\end{equation}
By Cauchy--Schwarz,
\begin{equation}
\sum_{t=1}^T
\|x_t\|_{(A_{a_t}^{(t)})^{-1}}
\le
\sqrt{
T\sum_{t=1}^T
\|x_t\|_{(A_{a_t}^{(t)})^{-1}}^2
}.
\end{equation}
It remains to control the quadratic sum.
For each fixed action $a$, whenever $a_t=a$,
\begin{equation}
A_a^{(t+1)}=A_a^{(t)}+x_tx_t^\top.
\end{equation}
By the matrix determinant lemma,
\begin{equation}
\det(A_a^{(t+1)})
=
\det(A_a^{(t)})
\Bigl(1+\|x_t\|_{(A_a^{(t)})^{-1}}^2\Bigr).
\end{equation}
Taking logs and summing over all rounds with $a_t=a$,
\begin{equation}
\sum_{t:\,a_t=a}
\log\!\Bigl(1+\|x_t\|_{(A_a^{(t)})^{-1}}^2\Bigr)
=
\log\frac{\det(A_a^{(T+1)})}{\det(\lambda I_d)}.
\end{equation}
Since $\lambda\ge L^2$ and $A_a^{(t)}\succeq \lambda I_d$,
\begin{equation}
\|x_t\|_{(A_a^{(t)})^{-1}}^2
\le
\frac{\|x_t\|_2^2}{\lambda}
\le
1.
\end{equation}
Using $\log(1+u)\ge u/2$ for $u\in[0,1]$, we obtain
\begin{equation}
\sum_{t:\,a_t=a}
\|x_t\|_{(A_a^{(t)})^{-1}}^2
\le
2\log\frac{\det(A_a^{(T+1)})}{\det(\lambda I_d)}.
\end{equation}
Summing over all $a\in\mathcal{A}$,
\begin{equation}
\sum_{t=1}^T
\|x_t\|_{(A_{a_t}^{(t)})^{-1}}^2
\le
2\sum_{a\in\mathcal{A}}
\log\frac{\det(A_a^{(T+1)})}{\det(\lambda I_d)}.
\end{equation}
Finally, for each action $a$,
\begin{equation}
\log\frac{\det(A_a^{(T+1)})}{\det(\lambda I_d)}
\le
d\log\!\left(1+\frac{TL^2}{\lambda d}\right),
\end{equation}
so
\begin{equation}
\sum_{t=1}^T
\|x_t\|_{(A_{a_t}^{(t)})^{-1}}^2
\le
2Kd\log\!\left(1+\frac{TL^2}{\lambda d}\right).
\end{equation}
Therefore,
\begin{equation}
R_T
\le
2\bar{\beta}_T(\delta)\,
\sqrt{
2KdT\log\!\left(1+\frac{TL^2}{\lambda d}\right)
},
\end{equation}
which proves the theorem.
\end{proof}

\paragraph{Remark.}
The main text uses Thompson Sampling in implementation.
Theorem~\ref{thm:regret} analyzes an optimistic instantiation built on the same action-wise linear reward estimator, in order to obtain a concise finite-time guarantee.

\section{Algorithm of \method }
\label{algo:omce}

In this section, we present the pseudocode for the \method algorithm.

The optional multi-attempt extension uses at most $P$ total attempts for each query, including the initial attempt, and stops early upon success.
After each failed attempt, an LLM extracts case-specific insights from the trajectory and feedback, records them in Episodic Memory, and injects them into the task prompt containing the original query using $\operatorname{Augment}$ to construct the next input. We denote the context for attempt $p$ by $\mathcal{C}_{\mathrm{aug},t}^{(p)}$, with $\mathcal{C}_{\mathrm{aug},t}^{(1)}=\mathcal{C}_{\mathrm{aug},t}$ defined in Eq.~\ref{eq:augmented_context}.
The loop below specifies how this within-task reuse interacts with memory refinement and controller learning.

\Needspace{5\baselineskip}
\noindent
\captionof{algorithm}{Online Meta-Control of Experience (\method): Inference and Learning Loop}
\label{alg:omce}

\begin{algorithmic}[1]
\STATE \textbf{Input:} Task query $q_t$; Episodic Memory
$\mathcal{M}_{\textsc{epi}}$; Guidance Memory
$\mathcal{M}_{\mathrm{guide}}$; bandit statistics
$\{A_a^{(t)},b_a^{(t)}\}_{a\in\mathcal A}$;
forgetting factor $\gamma$; maximum number of attempts $P$.
\STATE \textbf{Output:} Task trajectory $\tau_t$.
\STATE \emph{Convention:} $P$ includes the initial attempt
(thus $P-1$ retries).

\STATE \textcolor{gray}{// \textbf{Phase 1: Meta-Control (Retrieval \& Action Selection)}}
\STATE Retrieve candidates
$\mathcal{M}_{\mathrm{epi}}^{\mathrm{retr}}$ and
$\mathcal{M}_{\mathrm{guide}}^{\mathrm{retr}}$
using lightweight retrieval on $q_t$.
\STATE Construct meta-state $x_t$ from the retrieved candidates.
\STATE Estimate
$\hat\theta_a^{(t)}=(A_a^{(t)})^{-1}b_a^{(t)}$
for each $a\in\mathcal A$
(Eq.~\ref{eq:theta_estimation}).
\IF{using Thompson Sampling}
  \STATE Sample
  $\tilde\theta_a^{(t)}\sim
  \mathcal N(\hat\theta_a^{(t)},\Sigma_a^{(t)})$
  for each $a$
  (Eq.~\ref{eq:ThompsonSampling_impl}).
  \STATE \textbf{Meta-Decision:}
  $a_t=\arg\max_{a\in\mathcal A}
  x_t^\top\tilde\theta_a^{(t)}$
  (Eq.~\ref{eq:decide_action}).
\ELSIF{using LinUCB}
  \STATE \textbf{Meta-Decision:}
  $\displaystyle\begin{aligned}[t]
    a_t=\arg\max_{a\in\mathcal A}\bigl(
    &x_t^\top\hat\theta_a^{(t)}\\[-1pt]
    &+\alpha\sqrt{
      x_t^\top(A_a^{(t)})^{-1}x_t}
    \bigr).
  \end{aligned}$
\ENDIF
\STATE Initialize
$\mathcal{C}_{\mathrm{aug},t}^{(1)}\leftarrow q_t$
and
$\Psi_{\mathrm{injected}}\leftarrow\emptyset$.

\STATE \textcolor{gray}{// \textbf{Phase 2: Inject Experience According to the Selected Action}}
\IF{$a_t=a_{\textsc{DIRECT}}$}
  \STATE
  $\mathcal{C}_{\mathrm{aug},t}^{(1)}\leftarrow q_t$.
\ELSIF{$a_t=a_{\textsc{EPISODE}}$}
  \STATE
  $\mathcal{C}_{\mathrm{aug},t}^{(1)}
  \leftarrow
  \operatorname{Augment}
  (q_t,\mathcal{M}_{\mathrm{epi}}^{\mathrm{retr}})$.
\ELSIF{$a_t=a_{\textsc{GUIDE}}$}
  \STATE
  $\mathcal{C}_{\mathrm{aug},t}^{(1)}
  \leftarrow
  \operatorname{Augment}
  (q_t,\mathcal{M}_{\mathrm{guide}}^{\mathrm{retr}})$.
  \STATE
  $\Psi_{\mathrm{injected}}
  \leftarrow
  \mathcal{M}_{\mathrm{guide}}^{\mathrm{retr}}$.
\ELSIF{$a_t=a_{\textsc{DISTILL}}$}
  \STATE
  $\psi_{\mathrm{new}}
  \leftarrow
  \operatorname{DISTILL}
  (\mathcal{M}_{\mathrm{epi}}^{\mathrm{retr}})$
  (Eq.~\ref{eq:inductive_update}).
  \STATE
  $\mathcal{M}_{\mathrm{guide}}
  \leftarrow
  \mathcal{M}_{\mathrm{guide}}
  \cup\{\psi_{\mathrm{new}}\}$.
  \STATE
  $\mathcal{C}_{\mathrm{aug},t}^{(1)}
  \leftarrow
  \operatorname{Augment}
  (q_t,\{\psi_{\mathrm{new}}\})$.
  \STATE
  $\Psi_{\mathrm{injected}}
  \leftarrow\{\psi_{\mathrm{new}}\}$.
\ENDIF

\FOR{attempt $p=1,\ldots,P$}
  \STATE \textcolor{gray}{// \textbf{Phase 3: Agent Solves the Task}}
  \STATE The agent generates
  $\tau_t^{(p)}
  \sim\pi_{\mathrm{LLM}}
  (\cdot\mid\mathcal{C}_{\mathrm{aug},t}^{(p)})$.
  \STATE Observe immediate reward $r_t^{(p)}$.

  \STATE \textcolor{gray}{// \textbf{Phase 4: Record \& Reflection-Conditioned Scaling}}
  \IF{$r_t^{(p)}$ indicates success}
    \STATE Extract case-specific insights
    $\Omega_t=\{\omega_{t,m}\}_{m=1}^{M_t}$.
    \STATE Form
    $\eta_{\mathrm{new}}
    =(\tau_t^{(p)},r_t^{(p)},\Omega_t)$.
    \STATE Record:
    $\mathcal{M}_{\mathrm{epi}}
    \leftarrow\mathcal{M}_{\mathrm{epi}}
    \cup\{\eta_{\mathrm{new}}\}$.
    \STATE \textbf{break}
    \textcolor{gray}{// Early stop upon success}
  \ELSE
    \STATE Extract case-specific insights
    $\Omega_t^{(p)}$ from the failed trajectory
    using an LLM.
    \STATE Record:
    $\mathcal{M}_{\mathrm{epi}}
    \leftarrow\mathcal{M}_{\mathrm{epi}}
    \cup\{(\tau_t^{(p)},r_t^{(p)},\Omega_t^{(p)})\}$.
    \STATE
    $\mathcal{C}_{\mathrm{aug},t}^{(p+1)}
    \leftarrow
    \operatorname{Augment}(q_t,\Omega_t^{(p)})$.
  \ENDIF
\ENDFOR

\STATE \textcolor{gray}{// \textbf{Phase 5: Refine Injected Guidance}}
\IF{$\Psi_{\mathrm{injected}}\neq\emptyset$}
  \FORALL{$\psi\in\Psi_{\mathrm{injected}}$}
    \STATE Refine
    $\psi'\leftarrow
    \operatorname{Refine}
    (\psi,\tau_t^{(p)},r_t^{(p)})$
    using an LLM
    (Eq.~\ref{eq:deductive_update}).
    \STATE Update $\mathcal{M}_{\mathrm{guide}}$
    by replacing $\psi$ with $\psi'$.
  \ENDFOR
\ENDIF

\STATE \textbf{Meta-Evolution:}
Update the bandit reward model's sufficient statistics:
\STATE
$\displaystyle\begin{aligned}[t]
  A_{a_t}^{(t+1)}
  &=\gamma A_{a_t}^{(t)}
  +\mathbf{1}\{a_t=a\}x_tx_t^\top,\\
  b_{a_t}^{(t+1)}
  &=\gamma b_{a_t}^{(t)}
  +\mathbf{1}\{a_t=a\}r_t^{(p)}x_t.
\end{aligned}$
(Eq.~\ref{eq:stat_updated}).

\end{algorithmic}

\section{Details for Experimental Design}
\label{app:design}

\subsection{Task Setting.} Our experiments follow the online adaptation evaluation setting \cite{zhang2025ACE}, where each method processes the test stream in a single sequential pass. For every incoming instance, the agent first produces its prediction using the current policy, memory, or context, observes the task feedback, and then updates the corresponding memory or controller state before moving to the next instance. To make the comparison controlled, all methods use the same task order on each benchmark, the same backbone LLM, and the same evaluation pipeline. 
Each method is evaluated in a single run over each benchmark–backbone configuration due to the cost of LLM API calls. Reported success rates aggregate outcomes across benchmark instances. For methods supporting retries, each instance allows up to four total attempts, including the initial attempt; these are within-instance retries rather than independent experimental runs.
Methods that support multi-attempt inference use a shared budget of $P=4$ total attempts, following \Cref{app:metrics}, and \method may stop early once a task is solved.

\subsection{Benchmarks.} Under this setting, we consider four complementary benchmarks spanning code generation, repository-level software engineering, realistic multi-turn conversation, and advanced general reasoning. English is the language used for all benchmarks.

\emph{Code generation and software engineering.} \textbf{BigCodeBench} \cite{zhuo2024bigcodebench} is a function-level code generation benchmark emphasizing complex instructions and diverse function or tool calls, with unit-test-based automatic evaluation to reflect practical programming behavior. \textbf{SWE-bench Verified} \cite{jimenez2023swe} is a human-validated subset of SWE-bench designed to assess whether agents can resolve real-world GitHub issues in repository-scale codebases.

\emph{Multi-turn conversation.} \textbf{MultiChallenge} \cite{deshpande2025multichallenge} measures whether a model can handle realistic, challenging user interactions over extended dialogue contexts, with particular emphasis on instruction-following, context allocation, and in-context reasoning. We use GPT-4o-mini \cite{web:openai:gpt4omini} as the judge model, to judge whether the answer meets the hidden task requirement.

\emph{Advanced general reasoning.} \textbf{BIG-Bench Extra Hard (BBEH)} \cite{kazemi2025big} is a substantially more challenging successor to BIG-Bench Hard (BBH) \cite{srivastava2023beyond,suzgun2023challenging}, covering a broad range of general reasoning skills. We use a shuffled 6-task subset comprising 1,200 instances: causal\_understanding \cite{nie2024moca, kiciman2023causal}, dyck\_languages \cite{tyen2023llms}, temporal\_sequence, web\_of\_lies \cite{white2024livebench}, word\_sorting \cite{tyen2023llms}, and zebra\_puzzles \cite{shah2024causal}. These instances are mixed and randomly shuffled to construct the evaluation stream.

\subsection{Evaluation Metrics.}
\label{app:metrics}

We adopt \textbf{one-shot success rate} (\textbf{OSR}) as the primary evaluation metric, defined as the fraction of test instances solved correctly on the first attempt during online adaptation. This metric directly measures the agent's immediate response quality under its current policy and accumulated experience. 
When multiple attempts are allowed, we additionally report \textbf{cumulative success rate} (\textbf{CSR}), defined as the fraction of instances solved within a maximum budget of $P=4$ total attempts, including the initial attempt. This corresponds to at most three retries after the initial attempt.
For methods without multi-attempt inference, the unavailable metric is marked as ``-'' in the result tables. Since \textbf{Reflexion} \cite{shinn2023reflexion} improves later attempts through within-instance self-reflection rather than cross-task experience transfer, we report it under CSR in the main comparison.

To measure inference efficiency, we report \textbf{tokens per task}, the average total LLM tokens consumed per instance. This count includes all LLM calls made throughout the full online adaptation pipeline for a task, rather than only the final task-solving call; in particular, it includes token usage from auxiliary stages such as memory update and other adaptation-related inference, as well as additional token consumption introduced by test-time scaling in baselines that use it. The average is computed by dividing the total token consumption by the total number of tasks.
Token accounting includes the cost of solving tasks and the cost of extracting and refining experiences, but excludes evaluator or judge-model calls.

\subsection{Baselines and Methods.}

For baselines and compared methods, we include a Base LLM and four representative test-time improvement approaches that enhance performance without parameter updates. These baselines separate test-time scaling alone, cross-task experience transfer alone, and methods that combine cross-task adaptation with multi-attempt inference.

\textbf{Base LLM.} Base LLM is evaluated on each benchmark without any external memory, i.e., the model is queried with the benchmark’s default prompt template (as provided by the dataset authors) and produces an answer directly.

\textbf{Reflexion \cite{shinn2023reflexion}.} Reflexion instantiates test-time scaling through within-instance self-reflection over repeated attempts. After an attempt, the agent converts task feedback into a natural-language self-reflection and conditions subsequent attempts on these reflections. In our comparison, Reflexion does not transfer experience across different test instances.

\textbf{MemP \cite{fang2025memp}.} MemP transfers experience across tasks but makes only a single attempt per instance. It therefore represents cross-task experience transfer without multi-attempt inference. 

\textbf{ACE \cite{zhang2025ACE}.} ACE treats prompts/contexts as evolving playbooks and improves them through a modular generate-reflect-curate loop with structured, incremental updates designed to prevent context degradation across iterations; ACE supports both offline prompt optimization and online adaptation via memory/context updates driven by execution feedback.

\textbf{ReasoningBank \cite{ouyang2025reasoningbank}.} We evaluate ReasoningBank in its sequential MaTTS setting, where the method combines cross-task adaptation with multi-attempt inference. 

\textbf{\method (ours).} \method combines memory evolution with online meta-evolution. \method-TS uses Thompson Sampling \cite{agrawal2013thompson}, while \method-UCB uses an optimistic LinUCB-style rule \cite{chu2011contextual,abbasi2011improved}. Appendix~\ref{app:action_selection} gives the formal specifications, and Appendix~\ref{app:theorem} analyzes the optimistic variant.

\paragraph{Controller and retrieval configuration.}
The shared configuration used by both variants is listed in Table~\ref{tab:controller_hyperparameters}. The retrieval count, state dimension, forgetting factor, regularization parameter, and exploration parameters were kept fixed across all benchmarks and backbone models.

\begin{table}[t]
\centering
\setlength{\tabcolsep}{5pt}
\renewcommand{\arraystretch}{1.1}
\begin{tabular}{ll}
\toprule
\textbf{Quantity} & \textbf{Value} \\
\midrule
Retrieval count $M$ & 3 \\
Meta-state dimension $d$ & 9 \\
Forgetting factor $\gamma$ & 0.98 \\
Ridge regularization $\lambda$ & 0.5 \\
LinUCB coefficient $\alpha$ & 3.0 \\
TS posterior scale $v^2$ & 0.25 \\
Maximum total attempts $P$ & 4 \\
Retries after the initial attempt & 3 \\
\bottomrule
\end{tabular}
\caption{Shared controller and retrieval configuration used by \method-TS and \method-UCB.}
\label{tab:controller_hyperparameters}
\end{table}

We selected the shared configuration once using a randomly sampled subset of 100 tasks from the 1,140-task BigCodeBench stream. The configuration was then fixed for the complete BigCodeBench stream and for all other benchmarks and backbone models. No additional hyperparameter tuning was performed on SWE-bench Verified, MultiChallenge, BBEH, or individual backbone models. The 100 selected BigCodeBench tasks remain part of the final evaluation stream, so the BigCodeBench result contains a limited overlap between hyperparameter selection and evaluation.

The four evaluation streams contain 1,140, 500, 273, and 1,200 instances for BigCodeBench, SWE-bench Verified, MultiChallenge, and BBEH, respectively. All methods use the same task order, backbone model, and evaluation pipeline for a given benchmark.

We use the same backbone LLM for task execution and for insight extraction, the DISTILL action, and guidance refinement in memory evolution.

\textbf{Common Implementation Details.} Across all evaluated methods, we enforce a consistent online adaptation setting where any memory, playbook, or experience library is initialized as empty at the beginning of the evaluation. For all retrieval operations, we measure the cosine similarity between the queries and memory entries, utilizing \texttt{text-embedding-3-large} \cite{openai2024embedding} to encode the text representations.

\paragraph{Benchmark implementations.}
We integrate the benchmark Python packages into a common evaluation framework. The public implementations are available in the repositories for BigCodeBench,\footnote{\url{https://github.com/bigcode-project/bigcodebench}} SWE-bench,\footnote{\url{https://github.com/SWE-bench/SWE-bench}} MultiChallenge,\footnote{\url{https://github.com/ekwinox117/multi-challenge}} and BBEH.\footnote{\url{https://github.com/google-deepmind/bbeh}}
We evaluate on the original 273-instance MultiChallenge dataset and the six-task BBEH subset described above, which contains 1,200 instances. For MultiChallenge, GPT-4o-mini serves as the judge model. All compared methods share the same benchmark data and evaluation pipeline.

\section{Detailed Efficiency Results}
\label{app:efficiency_details}

\begin{figure*}[htbp]
  \centering
  \subcaptionbox{GPT-5.2\label{fig:efficiency_gpt52}}{\includegraphics[width=0.98\linewidth]{Figures/performance_vs_efficiency_multi_panel_GPT-5.2.pdf}}
  \vspace{6pt}
  \subcaptionbox{Deepseek-V3.2\label{fig:efficiency_dsv32}}{\includegraphics[width=0.98\linewidth]{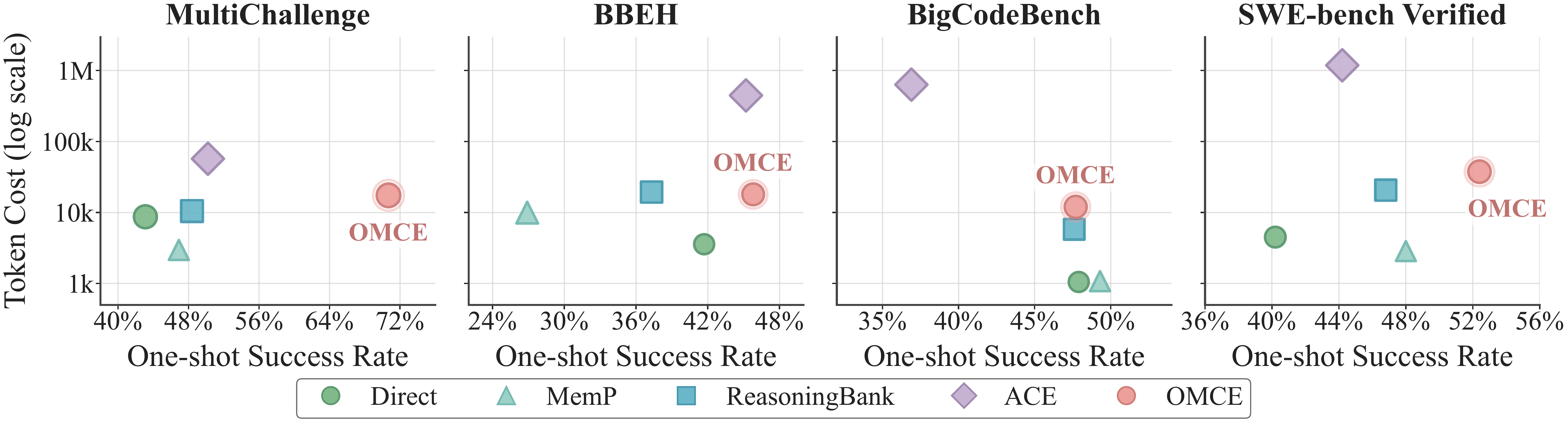}}
  \vspace{6pt}
  \subcaptionbox{Deepseek-V4-pro\label{fig:efficiency_dsv4pro}}{\includegraphics[width=0.98\linewidth]{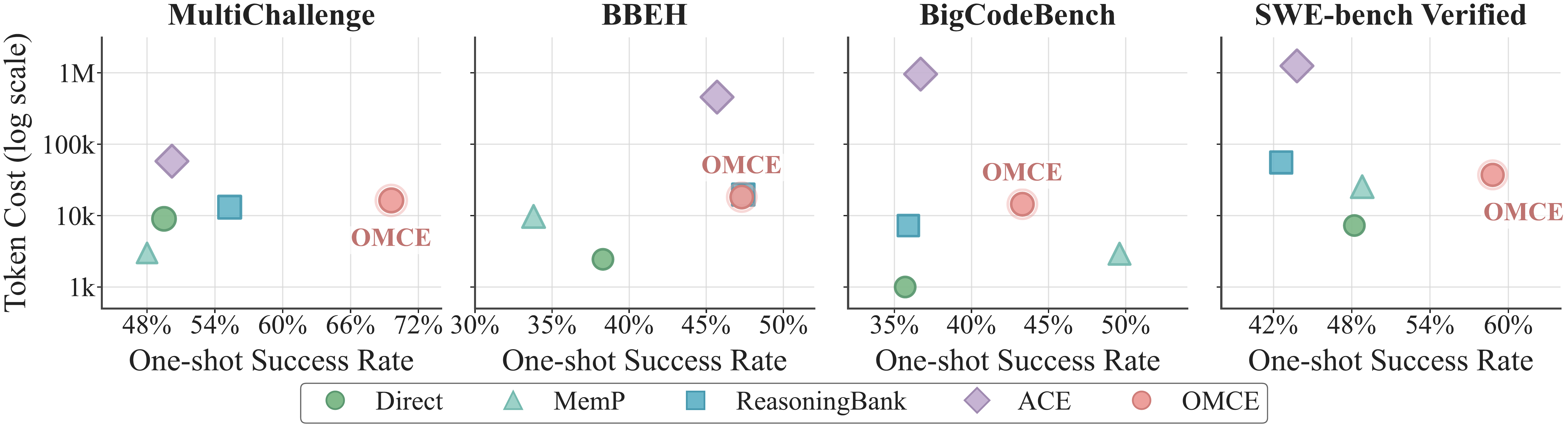}}
  \vspace{6pt}
  \subcaptionbox{Kimi-K2.5\label{fig:efficiency_kimi}}{\includegraphics[width=0.98\linewidth]{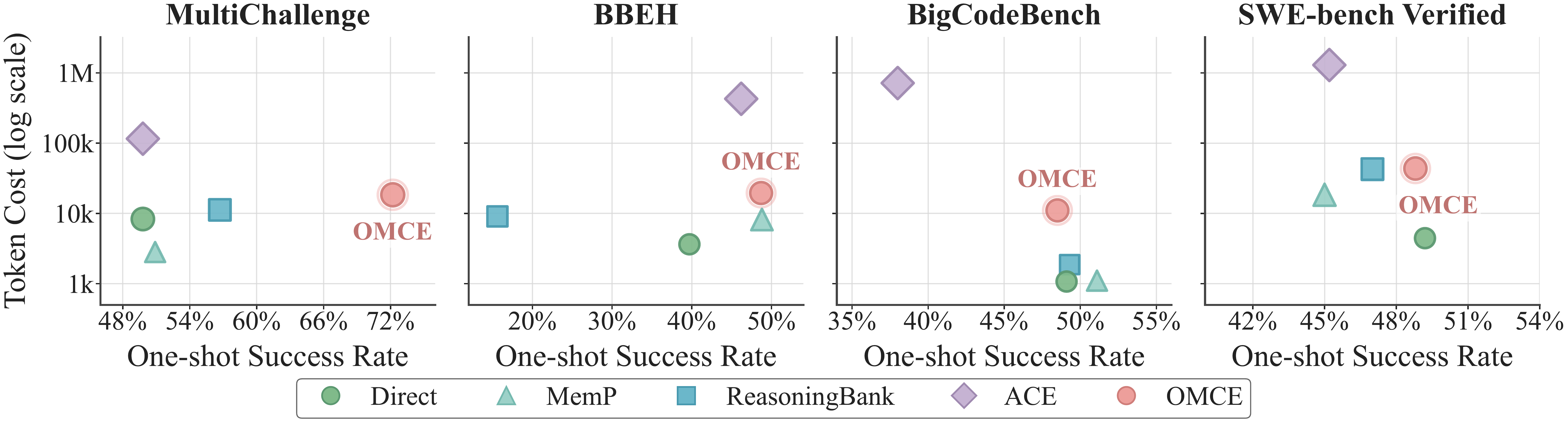}}
  \caption{Performance--efficiency trade-off on four benchmarks using GPT-5.2, Deepseek-V3.2, Deepseek-V4-pro, and Kimi-K2.5 as backbone LLMs. The x-axis shows one-shot success rate, and the y-axis shows token cost (log scale), defined as the average total number of prompt and output tokens consumed per task.}
  \label{fig:efficiency_all_backbones}
\end{figure*}

Figure~\ref{fig:efficiency_all_backbones} reports \textit{tokens per task}, as defined in \Cref{app:metrics}, across four backbone LLMs (GPT-5.2, Deepseek-V3.2, Deepseek-V4-pro, and Kimi-K2.5) on four benchmarks. Lower values indicate better inference efficiency.

The figure reveals a clear and consistent pattern across all four backbone models. Base LLM and MemP are the most token-efficient methods because they eschew test-time scaling, and therefore add limited overhead beyond direct inference. ReasoningBank and \method incur a moderate additional token budget due to retrieval, reflection, and structured online adaptation, but remain far more efficient than large-context prompting.

By contrast, ACE is dramatically less efficient: relative to \method, it requires about 62.5$\times$ more tokens per task on BigCodeBench, 32.5$\times$ more on SWE-bench Verified, 5.7$\times$ more on MultiChallenge, and 21.8$\times$ more on BBEH. This inefficiency is consistent with ACE's strategy of unconditionally injecting the entire playbook into the prompt, which substantially inflates the context length as adaptation proceeds.

Taken together with Figure~\ref{pic:performance_vs_efficiency}, these results place \method between the two extremes on every benchmark and backbone: more accurate than the token-cheap baselines, and far cheaper than ACE.

\section{Case Study on SWE-bench Verified}
\label{app:case_study}

This section provides a qualitative analysis of \method's behavior on the SWE-bench Verified dataset. Through three representative examples, we demonstrate how the Meta-Controller dynamically selects the optimal cognitive action---DIRECT, EPISODE, GUIDE, or DISTILL---to avoid harmful memory and improve the success rate of the underlying LLM.

\begin{table*}[t]
\centering
\small
\renewcommand{\arraystretch}{1.4} %
\resizebox{\textwidth}{!}{
\begin{tabular}{
  >{\raggedright\arraybackslash}p{0.18\linewidth} 
  >{\raggedright\arraybackslash}p{0.27\linewidth} 
  >{\raggedright\arraybackslash}p{0.26\linewidth} 
  >{\raggedright\arraybackslash}p{0.26\linewidth}
}

\toprule
\textbf{Pattern \& Action} & \textbf{Task Context} & \textbf{Baseline / Wrong Memory} & \textbf{\method Behavior \& Outcome} \\
\midrule

\textbf{Memory Harmful} \par
\vspace{2pt}
\colorbox{gray!15}{$a_{\textsc{DIRECT}}$} &
\textbf{ID:} \texttt{django-16560} \par
\textbf{Objective:} Add \texttt{violation\_error\_code} to \texttt{BaseConstraint}. \par
\textbf{Challenge:} Prone to wrong-layer patch trap. & 
\textbf{$a_{\textsc{GUIDE}}$:} Misled by irrelevant guidance entries, applying patches to the wrong abstraction layer. \par
\vspace{4pt}
\textcolor{red!80!black}{\ding{55} Failure} & 
\textbf{Reasoning:} \textit{Accurately identifies low relevance (confidence 1.0) and avoids hallucinated guidance entries.} \par
\vspace{4pt}
\textcolor{green!60!black}{\ding{51} Success} \\

\midrule

\textbf{Type Selection} \par
\vspace{2pt}
\colorbox{gray!15}{$a_{\textsc{GUIDE}}$} &
\textbf{ID:} \texttt{django-13401} \par
\textbf{Objective:} Ensure abstract model fields do not compare as equal across different models. \par
\textbf{Challenge:} Requires adjusting \texttt{\_\_eq\_\_}, \texttt{\_\_hash\_\_}, and \texttt{\_\_lt\_\_} to consider \texttt{field.model} instead of just \texttt{creation\_counter}. & 
\textbf{$a_{\textsc{EPISODE}}$:} Fails to generalize from specific past case-specific insights, leading to incomplete or incorrect equality logic. \par
\vspace{4pt}
\textcolor{red!80!black}{\ding{55} Failure} & 
\textbf{Reasoning:} \textit{Recognizes that high-level guidance entries provides better architectural instruction for core equality semantics than case-specific insights.} \par
\vspace{4pt}
\textcolor{green!60!black}{\ding{51} Success} \\

\midrule

\textbf{DISTILL} \par
\vspace{2pt}
\colorbox{gray!15}{$a_{\textsc{DISTILL}}$} &
\textbf{ID:} \texttt{django-13809} \par
\textbf{Objective:} Add \texttt{--skip-checks} option to the \texttt{runserver} command. \par
\textbf{Challenge:} Requires correctly wiring the new option through the command's argument parser and execution flow. & 
\textbf{$a_{\textsc{DIRECT}}$:} Fails to identify the correct modification layer and misses necessary execution flow updates. \par
\vspace{4pt}
\textcolor{red!80!black}{\ding{55} Failure} & 
\textbf{Reasoning:} \textit{Finds highly similar case-specific insights but no directly applicable guidance entries; distills new guidance entries to identify the correct modification layer.} \par
\vspace{4pt}
\textcolor{green!60!black}{\ding{51} Success} \\

\bottomrule
\end{tabular}
}
\caption{Qualitative case studies demonstrating the Meta-Controller's dynamic action selection on SWE-bench Verified.}
\label{tab:case_study}
\end{table*}

\paragraph{Avoiding Harmful Experience (\texttt{django-16560}).}
In this task, the objective is to add a \texttt{violation\_error\_code} parameter to Django's \texttt{BaseConstraint}. When forced to use guidance entries, the model is misled by irrelevant rules (e.g., "Wrong-Layer Patch Trap") and applies patches to the wrong abstraction layer, resulting in failure. However, the Meta-Controller accurately calculates low retrieval scores for both episodic records and guidance entries, confidently selecting DIRECT ($a_{\textsc{DIRECT}}$) (confidence 1.0). By proceeding without hallucinated guidance, \method successfully solves the task, demonstrating its ability to protect the model from harmful or irrelevant memory.

\paragraph{Selecting the Right Experience Type (\texttt{django-13401}).}
This task requires modifying the equality semantics of abstract model fields so they do not compare as equal across different models. This is a highly sensitive task where EPISODE fails because specific past cases do not generalize well to core equality logic. Conversely, GUIDE succeeds. The Meta-Controller successfully identifies this distinction, routing the task to $a_{\textsc{GUIDE}}$. This highlights the critical importance of selecting the appropriate experience type for tasks that require architectural guidance rather than specific implementation details.

\paragraph{DISTILL (\texttt{django-13809}).}
The goal of this task is to add a \texttt{--skip-checks} option to the \texttt{runserver} command. The baseline model without experience fails to correctly wire the option through the execution flow. The Meta-Controller observes that while there are highly similar episodic records available, there are no directly applicable guidance entries. Consequently, it selects DISTILL ($a_{\textsc{DISTILL}}$) to create a new guidance entry from the retrieved episodic records. The newly distilled guidance entry correctly instructs the model on the appropriate modification layer, leading to a successful resolution. This showcases \method's capability to dynamically generate new guidance entries when existing experience is insufficient.

\section{Broader Impact}
\label{sec:broader_impact}

This work introduces Online Meta-Control of Experience (\method), a framework designed to improve the efficiency and reliability of self-evolving LLM agents. The ability to selectively ignore misleading or redundant past experiences helps mitigate the propagation of errors in memory-enhanced agent systems. This contributes to the development of more robust AI assistants. However, like all frameworks that enhance the autonomy and efficiency of AI agents, \method could potentially be repurposed for malicious activities. While \method offers a pathway toward more intelligent and resource-efficient agents, its deployment should be accompanied by continuous monitoring and safety protocols to ensure that its evolutionary capabilities remain aligned with ethical standards and human values.

\end{document}